\documentclass[letterpaper,twocolumn,10pt]{article}
\usepackage{usenix}
\usepackage{amsmath}
\usepackage{graphicx} 
\usepackage{subcaption}
\usepackage{multirow}

\usepackage{00_preamble}
\usepackage{00_macros}
\usepackage{00_spacing}

\begin{document}

\title{\LARGE \bf Probabilistic Segmentation for Robust Field of View Estimation}

\newif\ifAnonymize

\Anonymizefalse

\ifAnonymize

\else
    \author{R. Spencer Hallyburton, David Hunt, Yiwei He, Judy He, and Miroslav Pajic%
    \thanks{This work is sponsored in part by the ONR under agreement N00014-23-1-2206, AFOSR under the award number FA9550-19-1-0169, and by the NSF under NAIAD Award 2332744 as well as the National AI Institute for Edge Computing Leveraging Next Generation Wireless Networks, Grant CNS-2112562.}
    \thanks{R. S. Hallyburton, D. Hunt, Y. He, J. He, and M. Pajic are with Department of Electrical and Computer Engineering, Duke University, Durham, NC 27708, USA;
    {\tt\small \{spencer.hallyburton,~miroslav.pajic\}@duke.edu}.}%
    }
\fi

\maketitle




\begin{abstract}
Attacks on sensing and perception threaten the safe deployment of autonomous vehicles (AVs). Security-aware sensor fusion helps mitigate threats but requires accurate field of view (FOV) estimation which has not been evaluated autonomy. To address this gap, we adapt classical computer graphics algorithms to develop the first autonomy-relevant FOV estimators and create the first datasets with ground truth FOV labels. Unfortunately, we find that these approaches are themselves highly vulnerable to attacks on sensing. To improve robustness of FOV estimation against attacks, we propose a learning-based segmentation model that captures FOV features, integrates Monte Carlo dropout (MCD) for uncertainty quantification, and performs anomaly detection on confidence maps. We illustrate through comprehensive evaluations attack resistance and strong generalization across environments. Architecture trade studies demonstrate the model is feasible for real-time deployment in multiple applications.
\end{abstract}
\section{Introduction}

Autonomous vehicles (AVs) rely on trusted situational awareness (SA) to navigate dynamic environments safely. However, AVs were not designed with security as a priority, making them susceptible to attacks on fundamental SA-building tasks such as perception and tracking. Recent research has demonstrated that attackers can exploit vulnerabilities in physical sensing channels~\cite{2019cao-spoofing, 2020sun-spoofing, 2022hally-frustum, petit2015remote,hunt_ndss24} and compromise systems at the cyber level~\cite{quinonez2020savior, hallyburton2023partial}. Such attacks manipulate SA by injecting false objects (false positives, FPs), removing real objects (false negatives, FNs), or altering object positions, leading to severe consequences for passengers and infrastructure.

To mitigate these threats, security-aware sensor fusion strategies execute consistency checks between expected and observed sensor data to flag anomalies. The two dominant approaches include: (1) a \emph{single-agent} framework, where detected objects are compared against historical predictions, and (2) a \emph{multi-agent} framework, where detected objects are validated via cross-agent comparisons.

Both approaches to security-aware fusion require accurate predictions of what an agent is \emph{expected} to see. These predictions depend on the visibility constraints of each sensor, motivating the need for a dynamic estimate of each sensor's \emph{field of view} (FOV) in real-time. Despite its critical role in AV security, FOV estimation has received limited attention due to the lack of focus on security and the nascency of multi-agent collaborative sensing. To advance research in assured autonomy, we draw upon techniques from the computer graphics community and demonstrate the first application of ray tracing~\cite{cook1984distributed} and concave hull estimation~\cite{park2012newconcave} to estimate FOV from LiDAR point clouds in AVs navigating dense environments.

To evaluate the performance of FOV estimators, we build the first autonomy-relevant datasets with ground truth FOV labels. We propose a labeling pipeline that can augment existing LiDAR-based datasets with accurate FOV labels to support evaluation in novel environments. We apply this methodology to diverse indoor and outdoor scenarios, including scenes from the nuScenes dataset~\cite{caesar2020nuscenes}, AV datasets generated using the CARLA simulator~\cite{hallyburton2023datasets}, and a custom-built unmanned ground vehicle (UGV) navigating a complex indoor environment. Our FOV estimation algorithms run in post-processing, producing high-quality ground truth FOV data useful for FOV estimation in LiDAR, radar, and camera-based sensing.

We consider the estimators in contested environments as a part of security-aware sensor fusion. Unfortunately, we find they are significantly vulnerable in adversarial settings; even simple attacks, such as indiscriminate LiDAR spoofing, significantly degrade FOV quality. These vulnerabilities render the traditional algorithms unsuitable, as even a weak adversary can manipulate the perceived FOV, leading to erroneous SA and an inability to reliably reason about sensor visibility.

Traditional FOV estimation is unsurprisingly vulnerable because it fits a boundary to all assumed-benign points. Ray tracing and concave hull techniques do not learn any characteristics of sensors, making them easy targets for adversaries. To address this limitation, we propose a learning-based FOV estimator that captures \emph{generalized features} of sensor visibility. Our estimator is a deep neural network (DNN) that predicts probabilities over a sensor's visibility in 3D space from learned FOV representations, even in the presence of adversaries. We leverage the generated FOV datasets to train models across applications on benign and adversarial data. 

In safety-critical applications, explicit uncertainty quantification is essential for reliable decision-making in unknown or contested environments. To model uncertainty in FOV estimation, we integrate Monte Carlo dropout (MCD)~\cite{gal2016dropout} into our DNN model, enabling approximate Bayesian variational inference. By applying stochastic dropout at inference time, MCD generates a probability distribution over FOV predictions, providing a measure of confidence in each estimate. Training the model on both benign and adversarial datasets demonstrates strong robustness against known attacks. Furthermore, we observe that adversarial perturbations not included in training induce significantly higher uncertainty in MCD outputs, motivating a novel FOV anomaly detection framework. This approach enhances security-aware sensor fusion by detecting previously unseen attacks on perception, further strengthening autonomy in contested environments.

To evaluate the generalization and scalability of our approach in real-time cyber-physical systems (CPS), we conduct large-scale experiments across multiple datasets. We assess the transferability of our DNN model by training and testing on different datasets, demonstrating strong generalization across both benign and adversarial conditions. Additionally, we explore trade-offs between resolution, inference speed, and model complexity through a parametric analysis of network architecture. This study illustrates that DNN-based FOV estimation is both efficient and accurate, supporting its viability for deployment in autonomous systems.

\textbf{In summary, our key contributions are:}
\begin{itemize}
    \setlength{\itemsep}{-4pt}
    \item First implementation of FOV estimation for autonomy using ray tracing and concave hull algorithms.
    \item Creation of the first FOV datasets derived from state-of-the-art simulated and real-world perception data.
    \item Demonstration of vulnerabilities in traditional FOV models under adversarial attacks.
    \item Development of a robust, learning-based FOV estimation approach using segmentation neural networks.
    \item Integration of Monte Carlo dropout for uncertainty-aware FOV estimation and anomaly detection.
\end{itemize}
\section{Autonomy in Contested Environments} \label{sec:system}

We describe algorithms that build SA, present critical adversarial threats against LiDAR sensing, and discuss security-aware sensing solutions that rely on accurate FOV estimates.

\subsection{Building Situational Awareness in AVs}

Sensors including LiDAR, cameras, and radar yield data on an environment that perception algorithms use to detect and track objects over time. Such tasks build longitudinal representations of scenes to help AVs plan safe paths through dynamic environments.

\subsection{Threats Against Sensor Fusion}

Despite significant advancements, perception and tracking remain vulnerable to occlusions, sensor noise, and adversaries that degrade SA. We consider attacks that compromise LiDAR-based perception as LiDAR are the most prevalent sensors in AVs for safety-critical 3D object detection and are intimately related to the FOV estimation task.

LiDAR sensors construct 3D representations of the environment but are vulnerable to attacks including, \textbf{LiDAR spoofing} where attackers inject laser pulses, creating false/negating objects in AV perception~\cite{2019cao-spoofing, 2020sun-spoofing, 2022hally-frustum}, \textbf{LiDAR jamming} where high-intensity lasers saturate or blind LiDAR sensors~\cite{petit2015remote}, and \textbf{point cloud manipulation} where adversaries perturb raw point cloud data, misleading object detection~\cite{hallyburton2023partial}.

\subsection{Security-Aware Sensor Fusion}

This section summarizes security-aware fusion techniques for single-agent and multi-agent configurations. We briefly describe the reliance of these approaches on FOV estimation.

\subsubsection{Single-Agent Security-Aware Sensor Fusion}

Fusion of multiple sensors on a single platform adds redundancy and enhances perception robustness. Accurate FOV estimation is needed to compare results between sensors and prevent occluded objects from being classified as adversarial. Key advancements in single-agent security include, validation of sensor data against physical constraints~\cite{zhu2013variational}, cross-sensor validation~\cite{xiang2023multi}, and learning-based methods~\cite{khazraei2022learning}. 

\subsubsection{Multi-Agent Security-Aware Sensor Fusion}

Fusion of data across agents enhances security awareness under the assumption that not all agents are compromised consistently. SA between agents is primarily relevant in regions of mutual overlap. The space of overlapping visibility depends heavily on sharing accurate FOV estimates. Approaches to multi-agent security include trust-based methods~\cite{hallyburton2024bayesian} and federated anomaly detectors~\cite{chellapandi2023federated}.
\section{FOV Estimation and Datasets in Autonomy}

Security-aware sensor fusion relies on accurate FOV estimation, yet few practical models exist for autonomy. This section describes how classical computer graphics techniques can be applied to estimate FOV from point cloud data and introduces the first labeled FOV datasets to advance FOV research.


\subsection{Algorithms for FOV Estimation}

\begin{figure}[t]
    \centering
    \begin{subfigure}[b]{0.48\linewidth}
        \centering
        \includegraphics[trim={0cm 0cm 2cm 4.5cm},clip,fbox,width=0.95\linewidth]{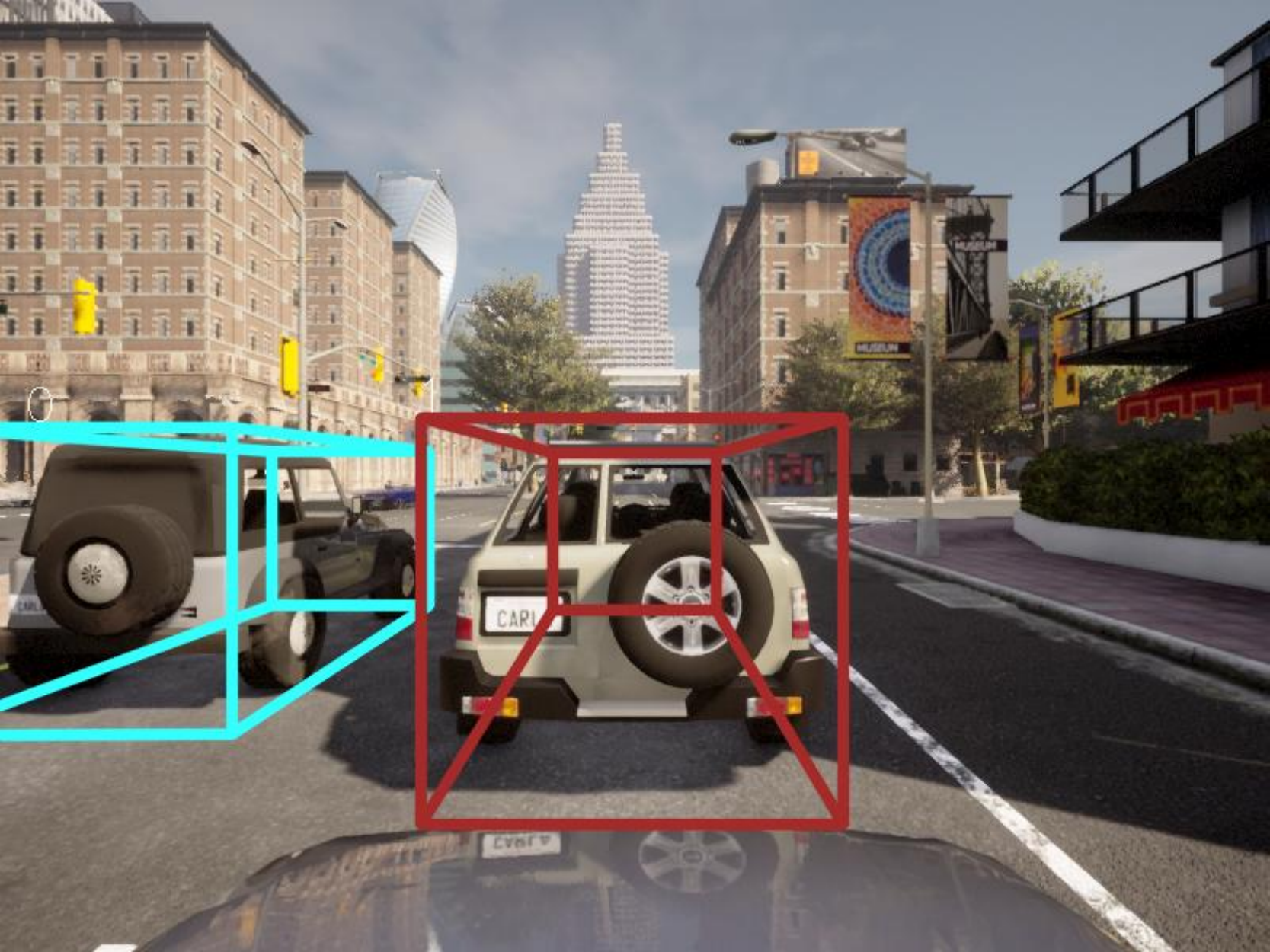}
        \caption{Ego facing occlusions.}
        \label{fig:system-carla-data-sample}
    \end{subfigure}
    %
    \begin{subfigure}[b]{0.48\linewidth}
        \centering
        \includegraphics[height=3cm,width=0.95\linewidth]{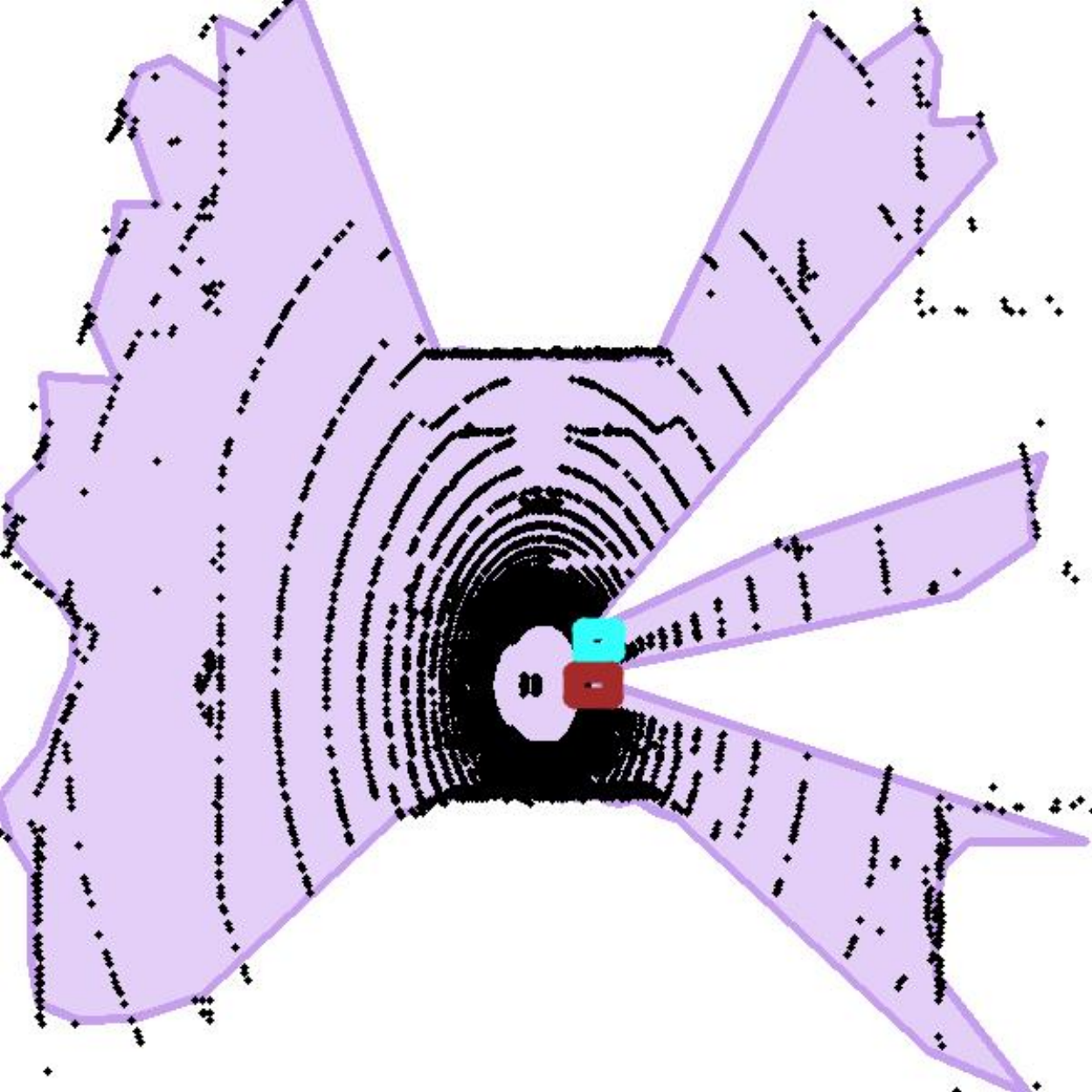}
        \caption{Visible region, raytrace FOV.}
        \label{fig:system-carla-data-fov}
    \end{subfigure}
    \caption{Space visible to the sensor is limited during occlusions from objects/infrastructure. Despite BEV LiDAR having a circularly symmetric FOV in wide open spaces, FOV is irregularly shaped in dense urban environments, and space behind nearby objects is not visible to the sensor. FOV algorithms fit model to visible space using point cloud as input and correctly captures lack of visibility behind objects.}
    \label{fig:system-carla-data}
\end{figure}

A sensor's FOV is highly dynamic, varying with agent position, sensor characteristics, and occlusions from static and dynamic objects. Existing security-aware fusion methods oversimplify FOV modeling, often assuming conical or circular visibility~\cite{golle2004detecting, soleymani2017secure, tsukada2022misbehavior, allig2019trustworthiness}. While an unimpeded LiDAR may exhibit a circular FOV, nearby objects significantly truncate visibility, as in Figure~\ref{fig:system-carla-data}. Some approaches incorporate static occlusions from buildings~\cite{bissmeyer2012assessment} but neglect dynamic obstacles, leading to erroneous visibility predictions and misalignment in sensor fusion.

To address these shortcomings, we adapt three classical computer graphics techniques to LiDAR-based FOV estimation. Each method takes as input LiDAR points projected into a bird’s-eye-view (BEV) reference. Examples of FOV estimation on BEV LiDAR data are in Figures~\ref{fig:system-carla-data} and~\ref{fig:dataset-samples}. 

\renewcommand{\subfigwidth}{0.30}
\begin{figure}
    \centering
    \begin{subfigure}[b]{\subfigwidth\linewidth}
        \includegraphics[width=\linewidth]{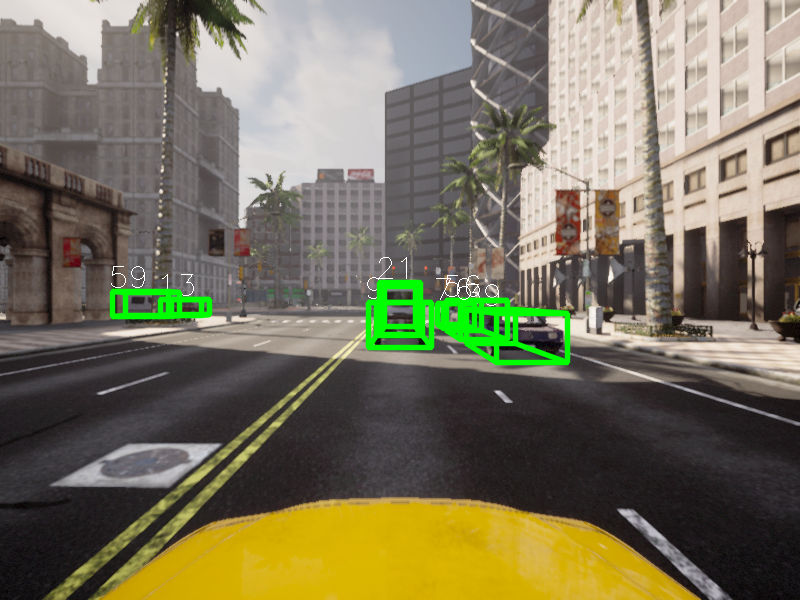}
        \caption{CARLA image}
        \label{fig:dataset-samples-carla-img}
    \end{subfigure}
    \begin{subfigure}[b]{\subfigwidth\linewidth}
        \includegraphics[trim={4.5cm 0 10cm 0},clip,width=\linewidth]{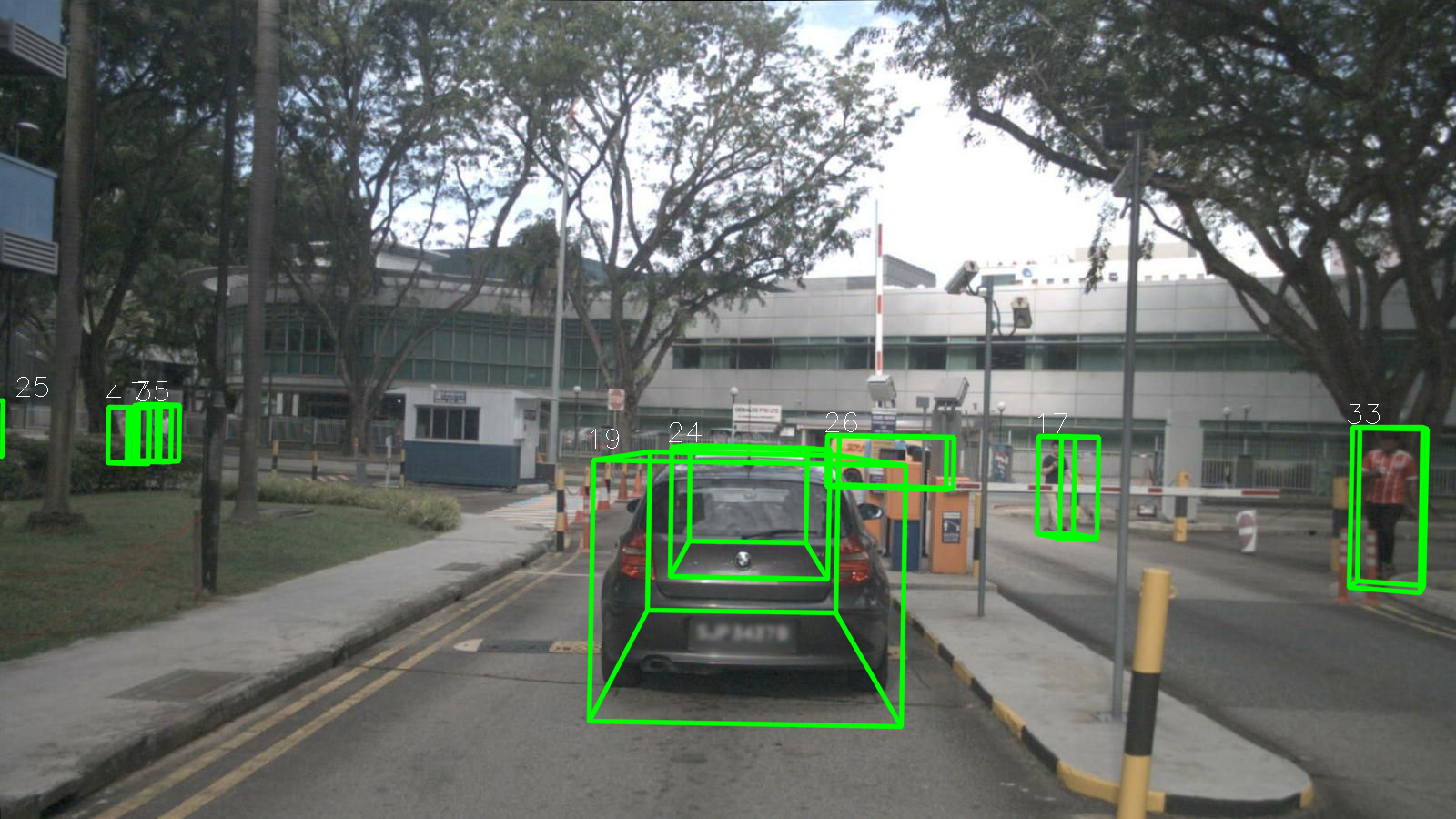}
        \caption{nuScenes image}
        \label{fig:dataset-samples-nuscenes-img}
    \end{subfigure}
    \begin{subfigure}[b]{\subfigwidth\linewidth}
        \includegraphics[trim={0cm 0 0.25cm 0},clip,width=\linewidth]{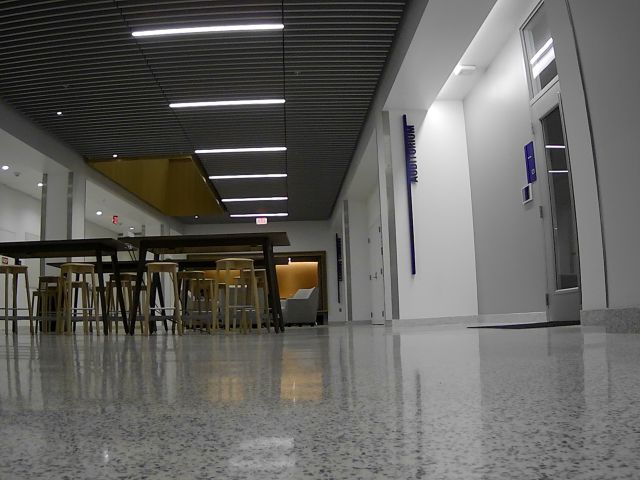}
        \caption{UGV image}
        \label{fig:dataset-samples-ugv-img}
    \end{subfigure}
    \begin{subfigure}[b]{\subfigwidth\linewidth}
        \includegraphics[width=\linewidth]{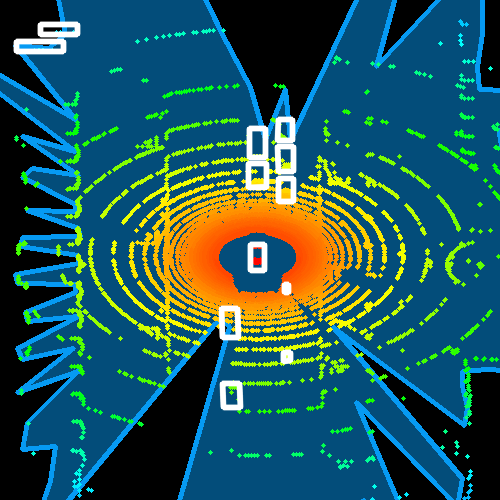}
        \caption{CARLA FOV}
        \label{fig:dataset-samples-carla-pc}
    \end{subfigure}
    \begin{subfigure}[b]{\subfigwidth\linewidth}
        \includegraphics[width=\linewidth]{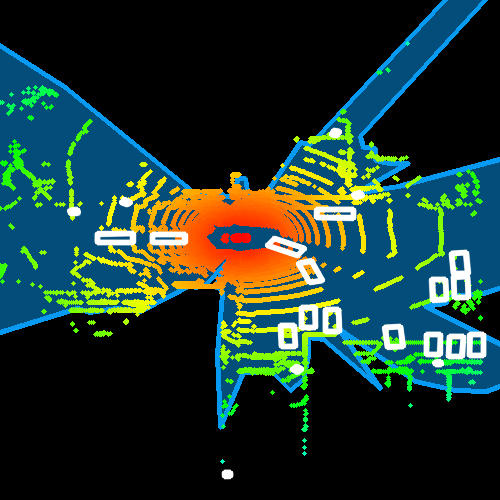}
        \caption{nuScenes FOV}
        \label{fig:dataset-samples-nuscenes-pc}
    \end{subfigure}
    \begin{subfigure}[b]{\subfigwidth\linewidth}
        \includegraphics[width=\linewidth]{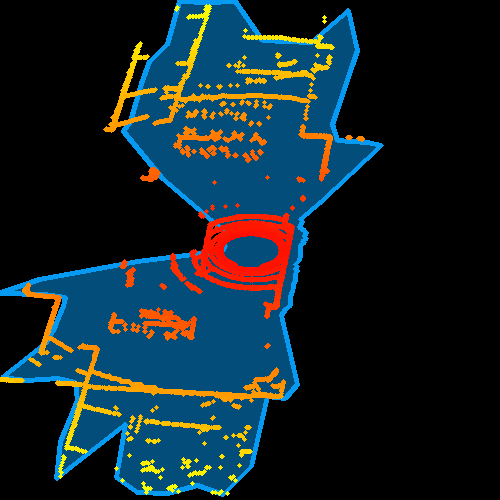}
        \caption{UGV FOV}
        \label{fig:dataset-samples-ugv-pc}
    \end{subfigure}
    \caption{BEV of LiDAR point clouds from state of the art simulators and real-world datasets support the first labeled FOV dataset. Scenes are diverse and capture indoor and outdoor environments with varying levels of occlusion.}
    \label{fig:dataset-samples}
\end{figure}

\noindent \textbf{Ray tracing} exploits the bijection between azimuth angles and LiDAR ranges in polar coordinates, ensuring points do not extend beyond physical barriers~\cite{cook1984distributed}. LiDAR returns are transformed into polar coordinates, mapping each azimuth angle to the maximum observed range.

\noindent \textbf{Concave hull estimation} fits a bounding polygon to LiDAR data without assuming a bijection, making it more flexible for occlusion-aware FOV estimation~\cite{park2012newconcave}. This method allows tunable parameters to control boundary tightness.

\noindent \textbf{Polygon fitting} refines 2D point sets, capturing complex non-convex regions with interior holes. PolyLidar~\cite{castagno2020polylidar3d} leverages Delaunay triangulation but is unsuitable for our application due to its reliance on dense point sets.


\subsection{Field of View Datasets}

While existing AV datasets emphasize object detection, they lack FOV annotations. We build the first labeled FOV dataset to enable benchmarking algorithms and advancing security-aware fusion. We discuss its composition in this section.

\begin{table}[t]

    
    \begin{subtable}{\linewidth}
    \centering
    \begin{tabular}{c||c|c|c}
    \textbf{CARLA} & Train & Validation & Test \\
    \toprule
    Scenes & 5 & 4 & 1 \\
    Agents & 10 & 10 & 10 \\
    Point clouds & 10,000 & 8,000 & 2000 \\
    Objects per agent & 6.5 & 5.1 & 7.2 \\
    \bottomrule
    \end{tabular}
    \end{subtable}


    \begin{subtable}{\linewidth}
    \centering
    \begin{tabular}{c||c|c|c}
    \textbf{nuScenes} & Train & Validation & Test \\
    \toprule
    Scenes & 700 & 150 & N/A \\
    Point clouds & 27,533 & 6,019 & N/A\\
    Objects per scene & 25.0 & 23.0 & N/A \\
    \bottomrule
    \end{tabular}
    \end{subtable}
    

    \begin{subtable}{\linewidth}
    \centering
    \begin{tabular}{c||c|c|c}
    \textbf{UGV} & Train & Validation & Test \\
    \toprule
    Scenes & 2 & 1 & N/A \\
    Point clouds & 8684 & 2372 & N/A\\
    Objects per scene & N/A & N/A  & N/A  \\
    \bottomrule
    \end{tabular}
    \end{subtable}

    \caption{We built first-of-a-kind FOV datasets derived from state-of-the-art datasets for perception in autonomy. Dataset is built from diverse outdoor and indoor environments for autonomous ground vehicles equipped with LiDAR sensing.}
    \label{tab:dataset-composition}
\end{table}

We construct FOV datasets from three sources: (1) the CARLA simulator~\cite{hallyburton2023datasets}, (2) the nuScenes dataset~\cite{caesar2020nuscenes}, and (3) a custom-built unmanned ground vehicle (UGV)~\cite{hunt2024radcloud}. Table~\ref{tab:dataset-composition} summarizes dataset composition, and Figure~\ref{fig:dataset-samples} provides representative samples. The datasets and code for generating FOV labels from CARLA simulations are released open-source.

\vspace{4pt}
\noindent \textbf{Carla dataset.} We generate longitudinal, multi-agent datasets in the CARLA simulator~\cite{dosovitskiy2017carla} using its expanded Python API~\cite{hallyburton2023datasets}. Multiple agents equipped with LiDAR and camera sensors operate alongside NPC vehicles, producing synchronized ground truth object locations and high-rate perception data. FOV estimation algorithms are applied to LiDAR point clouds, with a sample BEV output shown in Figure~\ref{fig:system-carla-data-fov}.

\vspace{4pt}
\noindent \textbf{nuScenes dataset.} nuScenes~\cite{caesar2020nuscenes} is a large-scale AV dataset containing 1,000 driving scenes, each spanning 20 seconds, captured in Boston and Singapore. It includes over 400K LiDAR sweeps and 1.4M object annotations across 23 categories, using sensors including a 32-beam LiDAR, 6 cameras, and 5 radars. We generate visibility maps in postprocessing.

\begin{figure}[!t]
    \centering
    \includegraphics[width=1.0\linewidth]{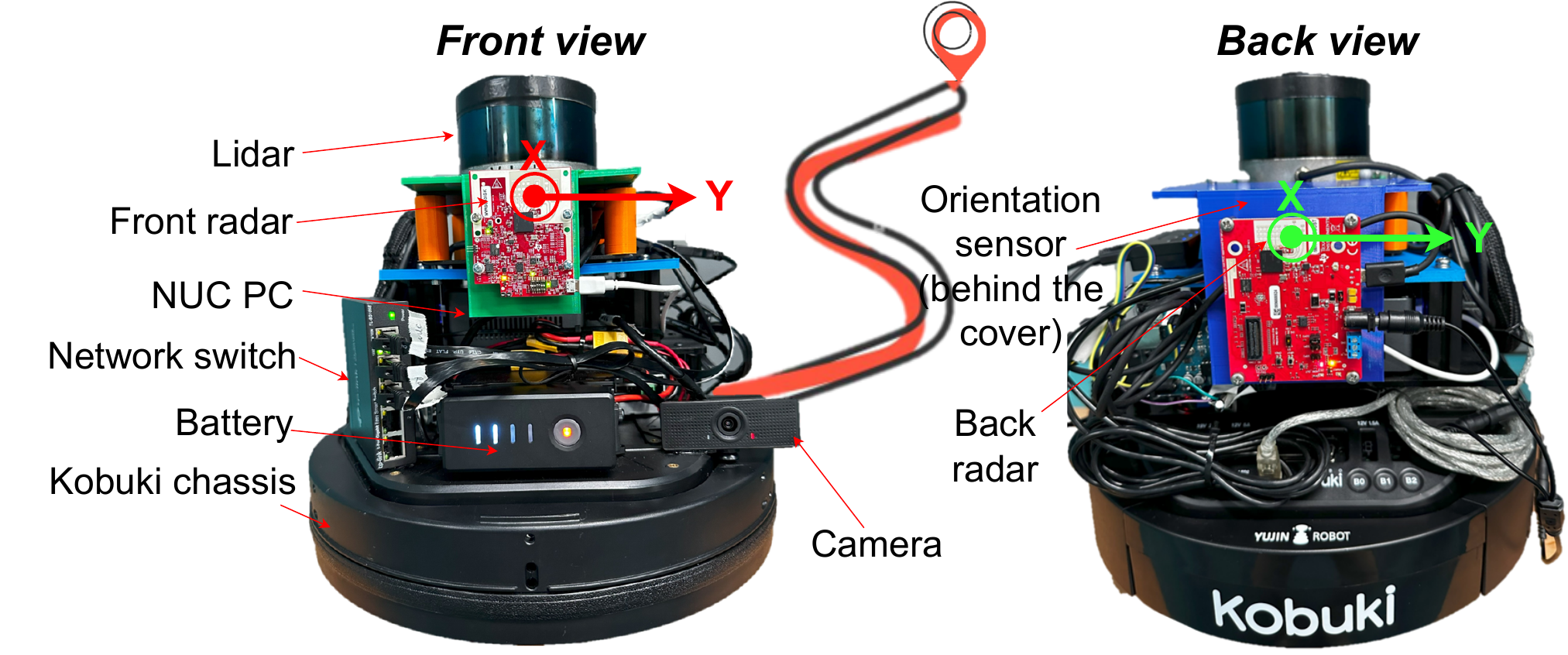}
\caption{The UGV maintains multiple sensors collecting time-synchronized data for perception and localization in dense, challenging indoor environments.}
\label{fig:experimental_platform}
\end{figure}

\vspace{4pt}
\noindent \textbf{UGV dataset.} We deploy a UGV equipped with a Velodyne VLP-16 LiDAR and a 2017 NUC7i5BNH CPU built on a Kobuki chassis as depicted in Figure~\ref{fig:experimental_platform}. The UGV captures data in environments with static and dynamic occlusions. With limited sensor range and mobility constraints, this dataset provides a challenging setting in highly obstructed conditions.

\section{Vulnerability Analysis of FOV Estimation}

We introduced ray tracing and concave hull FOV models to enhance security-aware sensor fusion in autonomy. Imperatively, for reliable fusion, the FOV algorithms must also be robust in contested environments. We evaluate the resilience of these classical techniques for FOV estimation and find they are broadly susceptible to even simple attacks on LiDAR.

\subsection{Threat Model}

Recently, LiDAR spoofing and cyber-based attacks have been demonstrated as practical and highly effective means of manipulating sensor data. These attacks inject synthetic points into the LiDAR sensing pipeline via physical or hardware channels, leading to false positive (FP) detections, false negative (FN) removals, and translated objects~\cite{2019cao-spoofing, 2020sun-spoofing, 2022hally-frustum, petit2015remote}.

We consider a conservative attack model where an adversary injects up to 150 spoofed points into the point cloud, a threshold well below the demonstrated capabilities of recent attacks~\cite{2019cao-spoofing, 2020sun-spoofing, 2022hally-frustum}. We consider two attack scenarios:
\begin{enumerate}
    \item \textbf{Cluster-Based Attack}: The adversary introduces dense clusters of points at specific locations to induce false detections~\cite{2020sun-spoofing,2022hally-frustum}.
    \item \textbf{Distributed Point Injection}: The attacker inserts points broadly across the LiDAR’s observable space without precise control to disrupt perception algorithms~\cite{petit2015remote,cao2021automated}.
\end{enumerate}

\subsection{Vulnerability of Traditional Algorithms}

\begin{figure}[t]
    \centering
    \begin{subfigure}[b]{0.62\linewidth}
        \centering
        \includegraphics[width=\linewidth]{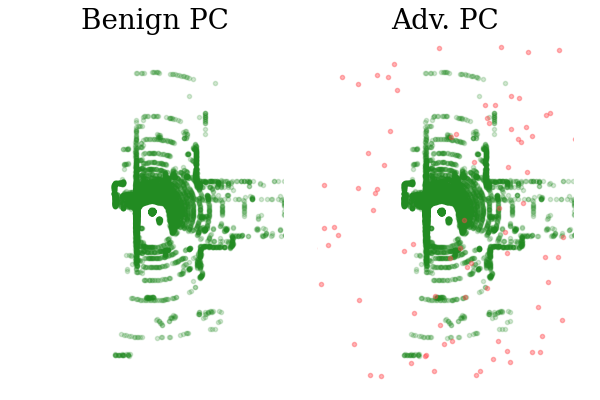}
        \caption{Point cloud in BEV frame.}
        \label{fig:vulnerability-fov-data}
    \end{subfigure}
    %
    \begin{subfigure}[b]{0.35\linewidth}
        \centering
        \includegraphics[width=\linewidth]{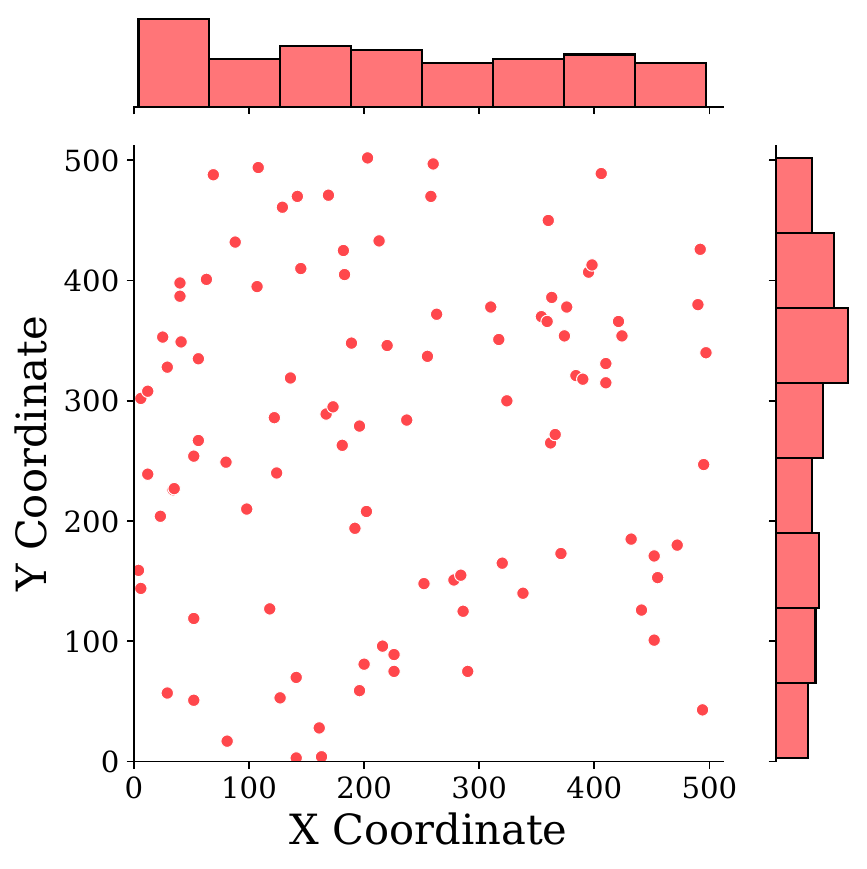}
        \caption{Spoof point spread.}
        \label{fig:vulnerability-fov-joint}
    \end{subfigure}
    %
    \begin{subfigure}[b]{\linewidth}
        \centering
        \includegraphics[trim={0cm 1.8cm 0cm 0cm},clip,width=\linewidth]{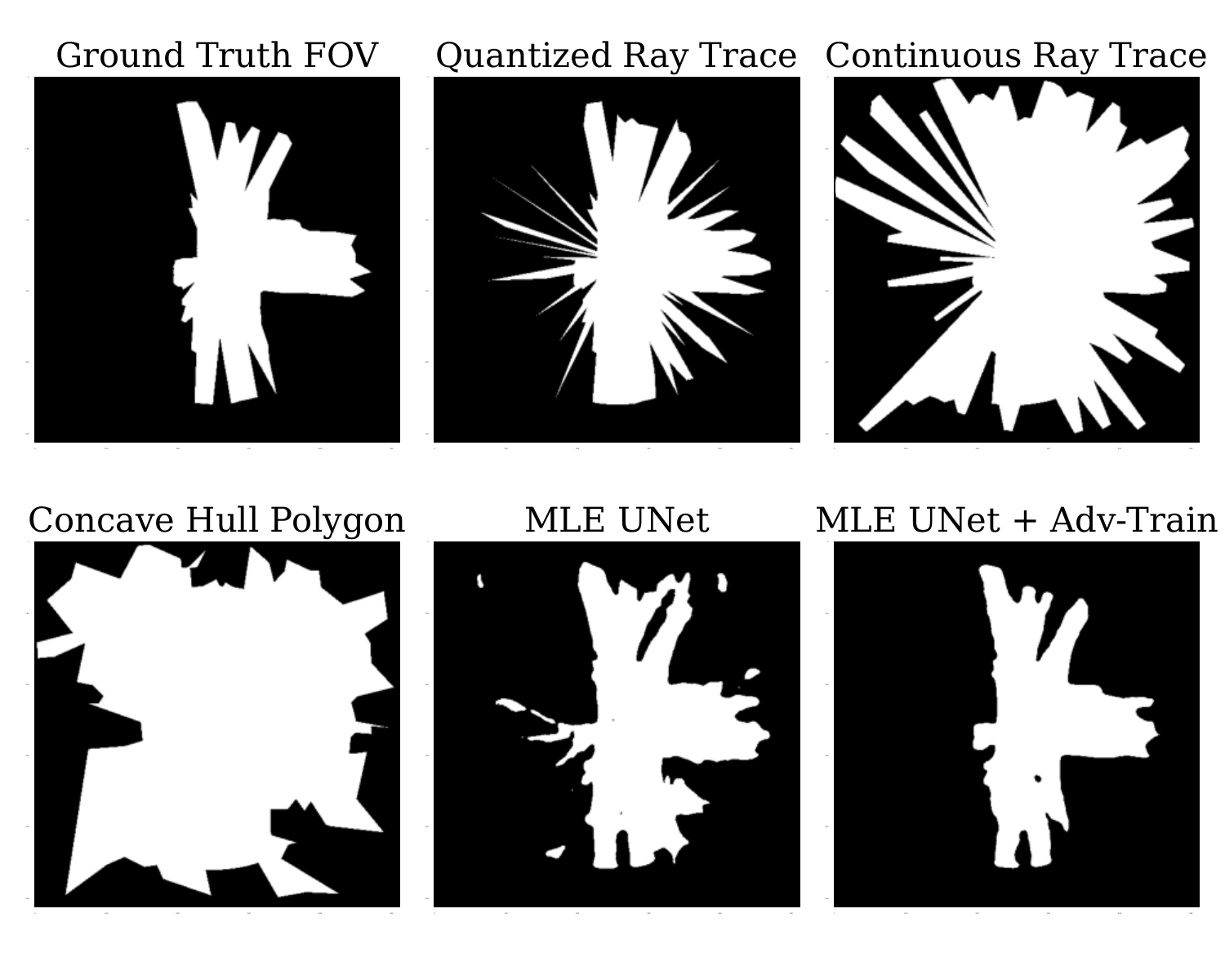}
        \caption{Traditional FOV algorithms vulnerable to uniform spoof, adv. training improves resiliency of UNet segmentation models.}
        \label{fig:vulnerability-fov-algorithms}
    \end{subfigure}
    \caption{FOV algorithms vulnerable to small number of spoof points - well under the demonstrated attacker capability from~\cite{2019cao-spoofing,2020sun-spoofing}. Adversarial training with UNet segmentation results in smoothed and robust FOV estimate.}
    \label{fig:vulnerability-fov}
\end{figure}

While previous LiDAR spoofing research focused on disrupting object detection, we find that these same attacks significantly degrade FOV estimation. We generate adversarial variants of the benign datasets by applying both cluster-based and distributed point injection attacks. Figure~\ref{fig:vulnerability-fov-data} illustrates these attacks on LiDAR point clouds. We then apply FOV estimation to the compromised data, revealing substantial distortions in the predicted FOV, as shown in Figure~\ref{fig:vulnerability-fov-algorithms}. 

Classical ray tracing and concave hull estimation assume all observed points contribute to the FOV boundary. These methods lack any ability to distinguish between authentic and adversarial data, as they do not incorporate learned priors or statistical features. Consequently, spoofed points appear indistinguishable from real LiDAR returns, leading to overestimated or malformed FOV predictions.

For instance, ray tracing, which relies on the azimuth-range mapping in polar coordinates, is particularly susceptible to distributed point injection. The presence of artificially injected points increases the estimated FOV, falsely suggesting that the sensor has visibility into occluded regions. Similarly, concave hull estimation, which fits a polygonal boundary to the LiDAR point cloud, expands the estimated FOV to enclose spoofed points, significantly distorting the sensor’s true visibility.

\subsection{Adaptive Attackers Overcome Defenses}

A straightforward approach to mitigating these vulnerabilities is to pre/post-process LiDAR point clouds to filter for adversarial points. Approaches include,
\begin{itemize}
    \item \textbf{Point Cloud Filtering}: Removing isolated or sparse points to eliminate spurious detections.
    \item \textbf{Clustering-Based Outlier Removal}: Identifying and discarding non-cohesive clusters that deviate from typical LiDAR return distributions.
    \item \textbf{Range Validation}: Filtering points beyond the sensor’s maximum range to prevent implausible spoofing.
\end{itemize}

However, while these approaches can reduce the impact of naive attacks, they are insufficient against an \emph{adaptive adversary}. Attackers can refine their spoofing strategies to evade detection, a phenomenon well-documented in adversarial machine learning research~\cite{athalye2018obfuscated}. By carefully crafting spoofed point distributions that resemble real object contours or leveraging time-coherent spoofing, an adaptive attacker can bypass simple filtering heuristics, preserving the effectiveness of the attack while evading traditional defenses.

Ultimately, the inability of classical FOV estimation algorithms to differentiate between benign and adversarial data highlights the need for learning-based models that incorporate statistical priors, adversarial training, and uncertainty.
\section{Approximately Bayesian Segmentation} \label{sec:segmentation}

We propose a learning-based approach to overcome the vulnerabilities of classical FOV algorithms. Our approach formulates FOV estimation as a binary segmentation task and employs a DNN designed to quantify uncertainty and improve adversarial robustness. The model takes as input LIDAR in a BEV reference and builds a probability map of visibility over a proximal region. The map is used to estimate FOV and detect anomalies. We describe each step in this section.

\subsection{Preprocessing}

Point clouds are inherently unordered, variable-sized, and permutation-invariant, properties that are incompatible with standard convolutional architectures. To enable structured grid-based processing, we apply a transformation that projects, filters, and quantizes the point cloud into a uniform single-channel matrix suitable for convolutional operations.

\subsubsection{Projection}

FOV estimation applies to both ground-vehicle AV sensors and infrastructure-mounted sensors with tilted vantage points. We project the 3D LiDAR point cloud into a bird's-eye-view (BEV) representation, accounting for the sensor's orientation using its quaternion attitude representation. This transformation simplifies the problem to a 2D grid while preserving spatial relationships relevant to ground-level object detection.

\subsubsection{Filtering}

We apply filtering to remove extraneous points irrelevant to practical FOV estimation. Points are discarded if they: (1) are unreasonably far from the sensor, or (2) are too high on the vertical channel (e.g.,~artifacts such as treetops or buildings). Filtering refines the dataset by ensuring that only points affecting perception-relevant visibility are retained.

\subsubsection{Quantization}

To map the BEV-projected points into a regular grid, we define a fixed spatial extent and output resolution. Each LiDAR return is assigned to a discrete grid cell using a transformation matrix similar to image rectification~\cite{hartley1999theory}. Cells accumulate the count of points mapped to them, forming a structured occupancy representation for input to the model.

\subsection{Model Design}

We perform binary segmentation on the quantized LiDAR data with a DNN model. We adopt a UNet architecture~\cite{ronneberger2015u} for its simplicity and mature implementation.

\subsubsection{Maximum Likelihood FOV Estimation}

\begin{figure}[t]
    \centering
    \includegraphics[width=\linewidth]{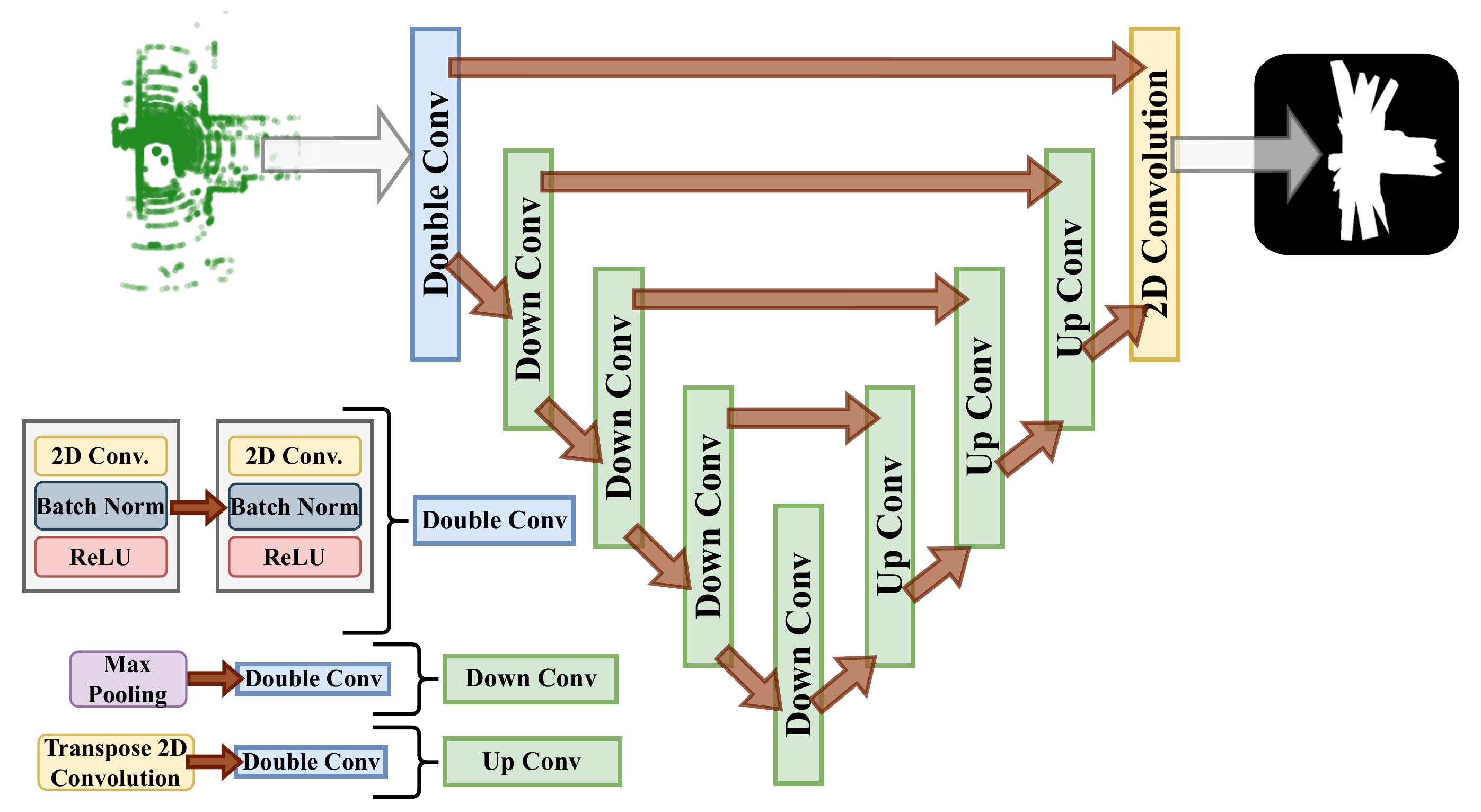}
    \caption{UNet architecture to ingest point cloud data mapped to BEV grid cells and output segmentation map. Utilizes a mixture of up and down convolutions with skip connections.}
    \label{fig:algorithm-architecture}
\end{figure}

Figure~\ref{fig:algorithm-architecture} depicts our network structure. An encoder first applies convolutional layers with max pooling to extract hierarchical features at progressively lower resolutions. The decoder then upsamples features while incorporating skip connections to preserve spatial information. The final layer outputs a probability map on $[0,1]$, where each grid cell's value represents the estimated likelihood of visibility. The network’s output is a maximum likelihood estimate (MLE) of the FOV. A probability threshold (e.g., $>0.7$) binarizes the prediction into (in)visible classes.

\subsubsection{Monte Carlo Dropout: Variational Approximation}

Standard MLE estimates do not capture model uncertainty, which is critical in adversarial settings where perturbed inputs can cause unpredictable behavior. To quantify uncertainty, we employ Monte Carlo dropout (MCD)~\cite{gal2016dropout} at test time, which enables a probabilistic interpretation of network predictions.

\begin{figure*}[t]
    \centering
    \begin{minipage}[c]{.2\linewidth}
    \begin{subfigure}[b]{\linewidth}
        \centering
        \includegraphics[width=0.95\linewidth,trim={0cm 0cm 0cm 0cm},clip]{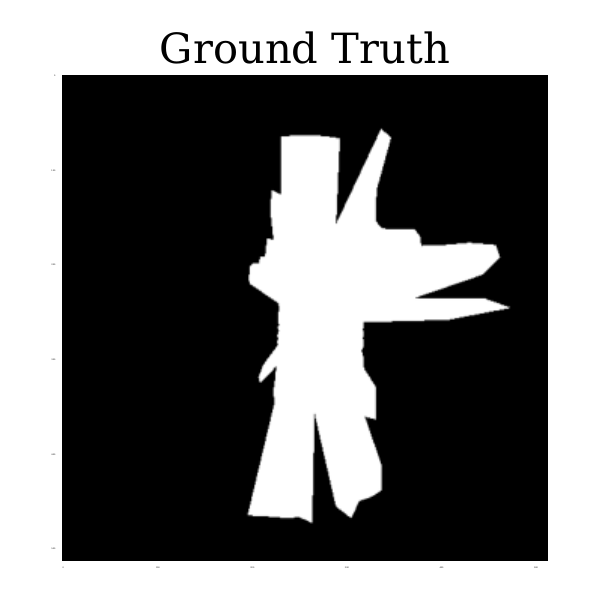}
        \caption{True FOV mask in BEV.}
        \label{fig:results-mcdropout-truth}
    \end{subfigure}
    \begin{subfigure}[b]{\linewidth}
        \centering
        \includegraphics[width=0.95\linewidth,trim={0cm 0cm 0cm 0cm},clip]{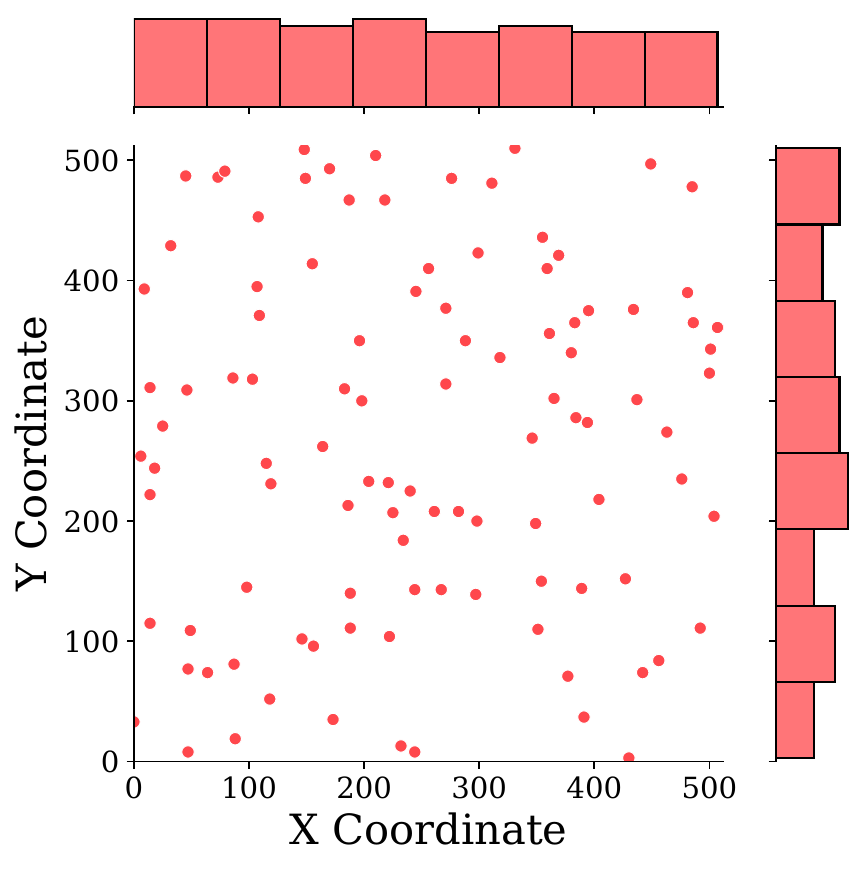}
        \caption{Distribution of small \# ``uniform spoof'' points.}
        \label{fig:results-mcdropout-spoof}
    \end{subfigure}
    %
    \end{minipage}\hfill
    \begin{minipage}[c]{.75\linewidth}
    %
    \begin{subfigure}[b]{\linewidth}
        \centering
        \includegraphics[width=0.95\linewidth,trim={0cm 0cm 0cm 0cm},clip]{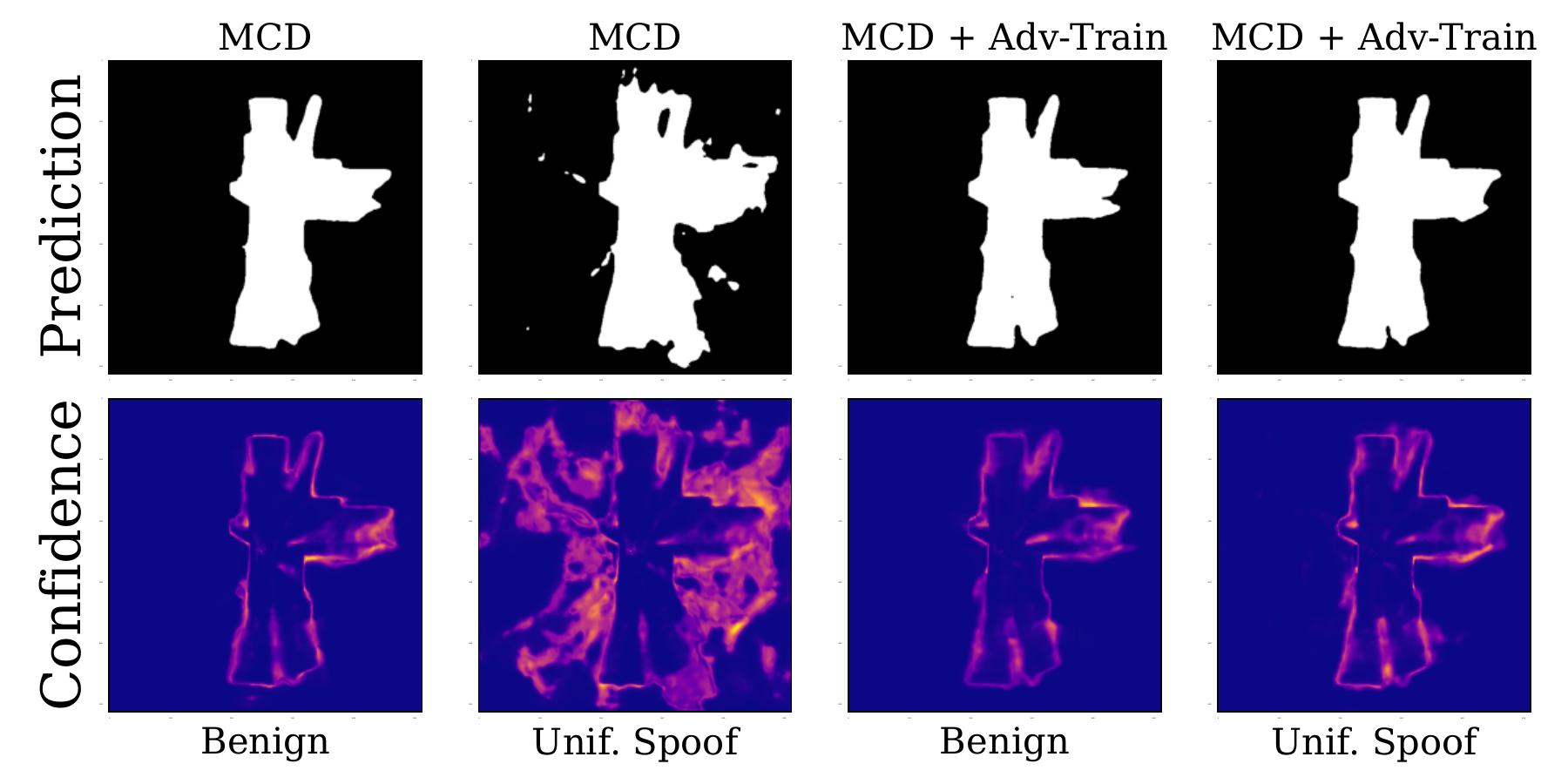}
        \caption{(col 1) MCD performs adequately in benign data without adv. training. (col 2) MCD (w/o adversarial training) displays high degree of uncertainty during spoof attack which can be used to detect attacks; however, output segmentation is compromised. (cols 3, 4) MCD with adv. training successfully determines FOV from both benign and adversarial inputs. (row 2, confidence) Brighter colors (red) represent less confidence/more uncertainty.}
        \label{fig:results-mcdropout-mcd}
    \end{subfigure}
    \end{minipage}
    \caption{A small number of spoofed points can compromise MCD UNet without adv. training. However, confidence map obtained from MC dropout is useful in detecting attacks due to large uncertainty. MCD with adv. training defends attack.}
    \label{fig:results-mcdropout}
\end{figure*}

Dropout is traditionally used during training to prevent overfitting~\cite{srivastava2014dropout}. However, MCD extends this by applying dropout during inference, approximating Bayesian variational inference~\cite{gal2016dropout}. By performing multiple forward passes with stochastic dropout, we obtain a probability distribution over the visibility of each grid cell. We compute the standard deviation ($1\sigma$) across predictions to generate a \textit{confidence map}, indicating regions of high uncertainty. Figure~\ref{fig:results-mcdropout-mcd} illustrates visually the outcomes of MCD applied to FOV estimation.

Alternative methods such as Bayesian convolutional networks~\cite{goan2020bayesian} offer more explicit probabilistic modeling but introduce additional complexity. We opt for MCD due to its simplicity and ease of integration into existing architectures.

\subsection{Integrity and Anomaly Detection}

Complementarily, we propose an anomaly detector centered on the MCD confidence maps that identifies grid cells where the variance of MCD visibility predictions exceeds a predefined threshold. Large uncertainty values suggest areas where the model is indecisive, potentially due to adversarial interference. By establishing a baseline uncertainty distribution from benign data, deviations can be flagged as potential attacks.

Upon detecting high uncertainty, the vehicle can perform security-aware actions such as cross-validate FOV estimates with redundant sensors (e.g., cameras, radar), monitor the LiDAR data for consistency over time, or adjust motion planning strategies to mitigate risk. This mechanism provides a proactive defense against adversarial manipulations, improving robustness in contested environments.

\subsection{Model Training} \label{sec:model-training}

We optimize hyperparameters of the network through 5-fold cross-validation, varying feature channel depth, dropout rate, and learning rate.  The final models are retrained with early stopping based on validation loss. Table~\ref{tab:algorithm-crossvalid} in Appendix~\ref{appendix:model-training} summarizes the hyperparameter search space and Figure~\ref{fig:algorithm-kfold} illustrates the training outcomes.

\section{Security Experiments} \label{sec:sec-experiments}

We conduct extensive evaluations of our trained models under both benign and adversarial datasets to assess their resilience against LiDAR spoofing attacks. 

\subsection{Large-Scale Model Evaluation} \label{sec:experiments-large-scale}

We systematically train and test our segmentation models across diverse datasets, evaluating their generalizability and robustness. To assess both in-distribution performance and cross-domain transferability, we train models on each dataset separately and test them across all dataset combinations.

\subsubsection{Training and Testing Protocol}

We train both MLE and MCD algorithms on benign and adversarial variants of the three primary datasets, yielding 12 trained models. Each model is trained for up to 30 epochs using batch sizes of 10 images on two NVIDIA A5000 GPUs. Hyperparameters, including dropout rate, feature channels, and learning rate, are selected during crossvalidation.

\begin{table*}[ht]
\centering
\begin{subtable}[c]{\linewidth}
\centering\begin{tabular}{lrrrrrrrrrrrr}
 & \multicolumn{4}{c}{Train: CARLA} & \multicolumn{4}{c}{Train: nuScenes} & \multicolumn{4}{c}{Train: UGV} \\
\cmidrule(l){2-5}\cmidrule(l){6-9}\cmidrule(l){10-13}
 & Prec. & Rec. & Acc. & F1 & Prec. & Rec. & Acc. & F1 & Prec. & Rec. & Acc. & F1 \\
\midrule
Test: CARLA & 0.95 & 0.90 & 0.97 & 0.92 & 0.91 & 0.92 & 0.96 & 0.91 & 0.95 & 0.83 & 0.95 & 0.88 \\
Test: nuScenes & 0.91 & 0.80 & 0.93 & 0.85 & 0.92 & 0.84 & 0.95 & 0.88 & 0.95 & 0.70 & 0.92 & 0.80 \\
Test: UGV & 0.92 & 0.80 & 0.95 & 0.85 & 0.91 & 0.84 & 0.96 & 0.87 & 0.97 & 0.79 & 0.96 & 0.87 \\
\bottomrule
\end{tabular}
\caption{Benign}
\label{tab:train-test-0}
\end{subtable}
\newline\vspace{6pt}\newline
%
%
%
\begin{subtable}[c]{\linewidth}
\centering\begin{tabular}{lrrrrrrrrrrrr}
 & \multicolumn{4}{c}{Train: CARLA Adv.} & \multicolumn{4}{c}{Train: nuScenes Adv.} & \multicolumn{4}{c}{Train: UGV Adv.} \\
\cmidrule(l){2-5}\cmidrule(l){6-9}\cmidrule(l){10-13}
 & Prec. & Rec. & Acc. & F1 & Prec. & Rec. & Acc. & F1 & Prec. & Rec. & Acc. & F1 \\
\midrule
Test: CARLA Adv. & 0.95 & 0.91 & 0.97 & 0.93 & 0.93 & 0.89 & 0.96 & 0.91 & 0.97 & 0.61 & 0.90 & 0.75 \\
Test: nuScenes Adv. & 0.92 & 0.76 & 0.93 & 0.83 & 0.93 & 0.77 & 0.93 & 0.84 & 0.97 & 0.55 & 0.89 & 0.69 \\
Test: UGV Adv. & 0.92 & 0.81 & 0.95 & 0.86 & 0.91 & 0.83 & 0.95 & 0.86 & 0.98 & 0.72 & 0.94 & 0.82 \\
\bottomrule
\end{tabular}
\caption{Train and test on adversarial}
\label{tab:train-test-2}
\end{subtable}
\newline\vspace{6pt}\newline
\begin{subtable}[c]{\linewidth}
\centering\begin{tabular}{lrrrrrrrrrrrr}
 & \multicolumn{4}{c}{Train: CARLA Adv. MC} & \multicolumn{4}{c}{Train: nuScenes Adv. MC} & \multicolumn{4}{c}{Train: UGV Adv. MC} \\
\cmidrule(l){2-5}\cmidrule(l){6-9}\cmidrule(l){10-13}
 & Prec. & Rec. & Acc. & F1 & Prec. & Rec. & Acc. & F1 & Prec. & Rec. & Acc. & F1 \\
\midrule
Test: CARLA Adv. & 0.97 & 0.85 & 0.96 & 0.90 & 0.93 & 0.85 & 0.95 & 0.89 & 0.95 & 0.71 & 0.92 & 0.81 \\
Test: nuScenes Adv. & 0.95 & 0.70 & 0.92 & 0.80 & 0.94 & 0.73 & 0.93 & 0.82 & 0.94 & 0.62 & 0.90 & 0.74 \\
Test: UGV Adv. & 0.93 & 0.76 & 0.95 & 0.84 & 0.91 & 0.78 & 0.95 & 0.84 & 0.95 & 0.79 & 0.95 & 0.86 \\
\bottomrule
\end{tabular}
\caption{Train and test on adversarial with MC dropout}
\label{tab:train-test-3}
\end{subtable}
\caption{We train models on benign and adversarial (adv.) data for each dataset and test the transferability of the models to new datasets without any fine-tuning. Models have high degrees of performance on precision, recall, accuracy, and F1-score metrics and generalize well to new datasets without any fine-tuning, even when transferring between indoor and outdoor environments.}
\label{tab:model-training}
\end{table*}

Each trained model is evaluated on a held-out test split, producing a binary mask predicting the (in)visibility of each grid cell in Cartesian space. Ground truth FOV masks from the datasets provide the reference for computing performance metrics, including precision, recall, F1-score, and area under the precision-recall curve (AUPRC). To assess generalization, each model is also tested on the unseen datasets, yielding a total of 72 combinations of models and test sets. We display a selection of the results in Table~\ref{tab:model-training}. We find that segmentation models are successfully able to generalize to new environments across indoor and outdoor domains with high degree of performance. The full set of results is available in the open-source evaluation notebooks, where pretrained models can be used to reproduce all findings.

\subsubsection{Security Analysis}

\begin{figure}[t]
    \centering
    \begin{subfigure}[b]{\linewidth}
        \includegraphics[width=0.75\linewidth]{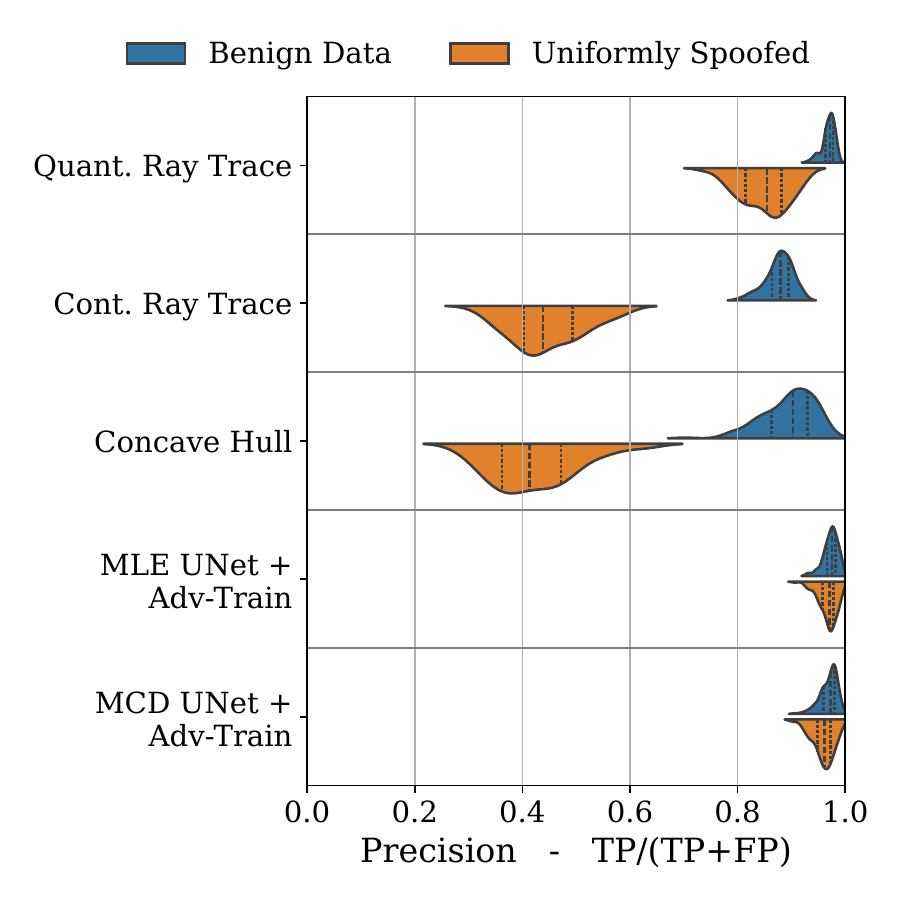}
    \end{subfigure}
    \begin{subfigure}[b]{\linewidth}
        \includegraphics[trim={0cm 0cm 0cm 1.25cm},clip,width=0.75\linewidth]{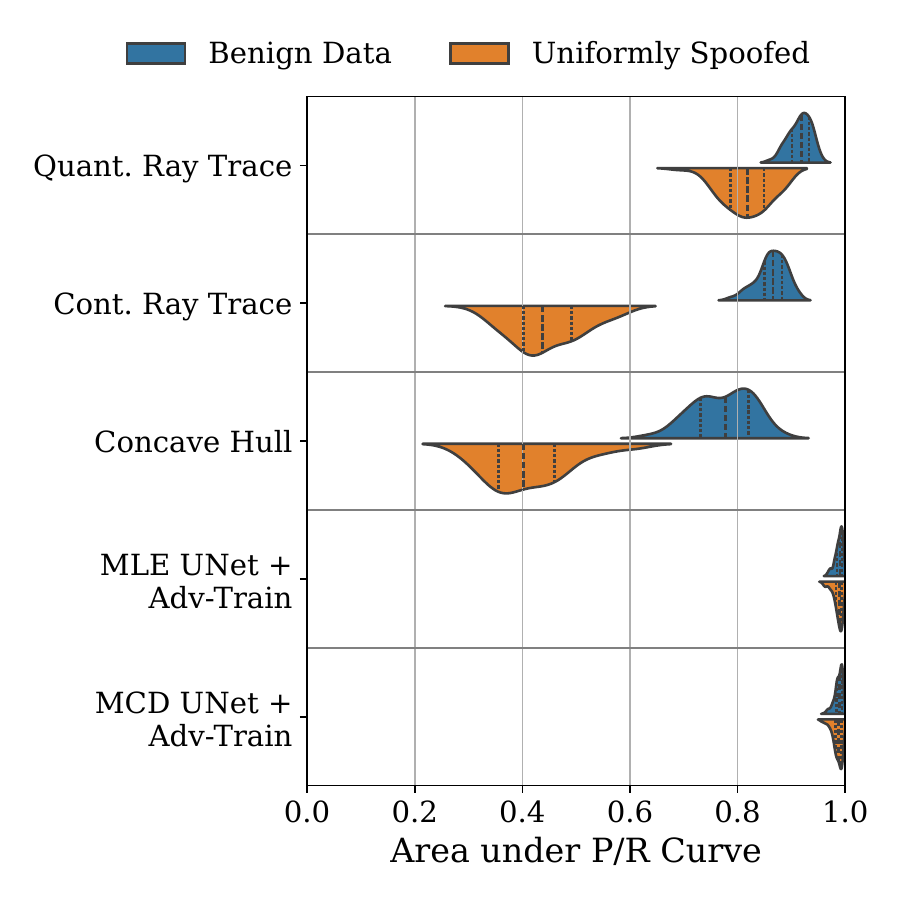}
    \end{subfigure}
    \caption{On testing set, model precision and AUPRC significantly deteriorate in the presence of spoofing attacks with even low numbers of spoofed points for all models except those with adversarial training. MCD network has benefit of providing confidence assessment on output FOV estimate.}
    \label{fig:results-violin}
\end{figure}

To illustrate the impact of adversarial spoofing attacks on different FOV estimation approaches, we summarize precision and AUPRC scores in Figure~\ref{fig:results-violin} for both classical and our MLE/MCD approaches. These metrics highlight the stark contrast in performance between traditional and learning-based models under benign and adversarial conditions.

\textbf{Vulnerability of traditional algorithms.} Classical FOV estimation methods including quantized ray tracing, continuous ray tracing, and concave hull estimation exhibit severe performance degradation under adversarial conditions. Spoofing attacks, even with as few as 100-150 injected points, cause precision losses ranging from $18\%$ to $60\%$, significantly distorting estimated FOV boundaries (see Figure~\ref{fig:vulnerability-fov}).

\textbf{Robustness of learning-based models.} In contrast, segmentation-based FOV estimators trained with adversarial samples exhibit minimal degradation. Models incorporating adversarial training maintain precision and recall scores nearly identical to those observed on benign data. MLE-based segmentation remains highly effective under attack, as illustrated visually in Figure~\ref{fig:results-mcdropout} and quantitatively in Table~\ref{tab:model-training} and Figure~\ref{fig:results-violin}. MCD-based models further provide uncertainty quantification, enhancing interpretability and security.



\subsection{Monte Carlo Dropout for Uncertainty}

Beyond its robustness to adversarial perturbations, MCD provides a probabilistic estimate of classification uncertainty. Figure~\ref{fig:results-mcdropout-mcd} compares MCD with and without adversarial training under both benign and spoofed LiDAR data. The adversarial dataset consists of uniformly distributed spoofed points in the X-Y plane (Figure~\ref{fig:results-mcdropout-spoof}). Key observations include,

\begin{itemize}
    \item \textbf{Resilience to known attacks.} MCD models trained with adversarial data maintain high classification accuracy even when tested on spoofed datasets.
    \item \textbf{Uncertainty as an anomaly indicator.} When an MCD model trained only on benign data encounters adversarial input, its confidence map exhibits significantly higher uncertainty values. This suggests that MCD can be leveraged for out-of-distribution (OOD) attack detection, identifying spoofed inputs even when they were not explicitly included in training.
\end{itemize}

These findings demonstrate that integrating uncertainty estimation into FOV models not only enhances resilience to known attacks but also provides a mechanism for detecting novel adversarial manipulations.

\section{Model Architecture Evaluation} \label{sec:arch-experiments}

The previous results illustrated that FOV estimation algorithms via DNN-based segmentation are robust to adversarial manipulations. The chosen architecture and hyperparameters in the security evaluation may not be suitable for all applications depending on requirements for accuracy, computational efficiency, and general robustness. 

Real-time platforms, such as UGVs/UAVs/AVs require fast inference with minimal latency, whereas high-precision mapping and infrastructure monitoring may prioritize accuracy over compute constraints. To ensure the versatility and adaptability of our FOV estimator, we systematically evaluate its performance across different model architectures and input resolutions, balancing accuracy, robustness, and efficiency.

Additionally, to show the potential for integration with full-stack AVs, we build FOV estimation into a near-real-time multi-agent robotics application using the Robot Operating System (ROS)~\cite{quigley2009ros}. Supplemental videos illustrating this performance are available online.

\subsection{Parametric Model Evaluation}

By systematically varying key model architectures parameters, we evaluate their impact on key metrics, including segmentation accuracy, adversarial robustness, inference latency, and computational load. 

\subsubsection{Considered Parameters}

To assess the trade-offs between accuracy and computational efficiency, we conduct a parametric evaluation over key architectural and input data parameters including,

\begin{itemize}
    \item \textbf{Model width.} Defined by the number of feature channels per layer, width affects the model’s ability to learn spatial representations. Wider models capture more complex features but require increased memory and computation. 
    \item \textbf{Model depth.} The number of layers in the network determines the  hierarchical feature extraction. Deeper networks may generalize better but can introduce diminishing returns, overfitting, and increased latency.
    \item \textbf{Input data resolution.} Higher-resolution inputs provide finer spatial granularity, improving segmentation accuracy but increasing inference time and memory usage.
\end{itemize}

\subsubsection{Performance Outcomes}

\renewcommand{\trimLR}{4cm}
\renewcommand{\trimBT}{2cm}
\renewcommand{\subfigwidth}{0.21}

\begin{figure*}[t]
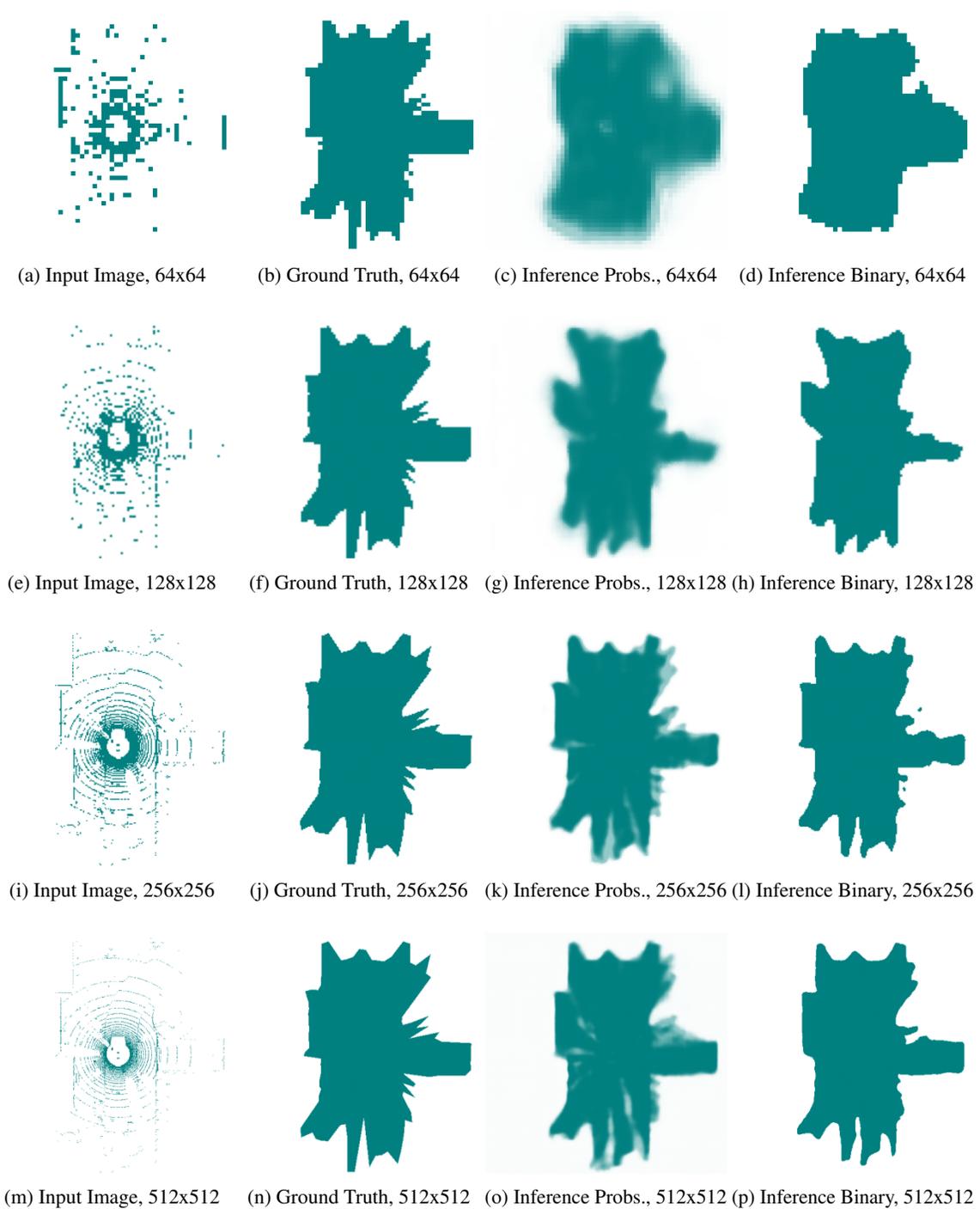

    \centering
    \foreach \x in {64,128,256,512}
    {
        \begin{subfigure}[b]{\subfigwidth\linewidth}
            \centering
            \includegraphics[trim={{\trimLR} {\trimBT} {\trimLR} {\trimBT}},clip,width=\linewidth]{diagrams/parametric-case/parametric_resol_\x_input_image.pdf}
            \caption{Input Image, {\x}x{\x}}
        \end{subfigure}
        %
        \begin{subfigure}[b]{\subfigwidth\linewidth}
            \centering
            \includegraphics[trim={{\trimLR} {\trimBT} {\trimLR} {\trimBT}},clip,width=\linewidth]{diagrams/parametric-case/parametric_resol_\x_gt_mask.pdf}
            \caption{Ground Truth, {\x}x{\x}}
        \end{subfigure}
        %
        \begin{subfigure}[b]{\subfigwidth\linewidth}
            \centering
            \includegraphics[trim={{\trimLR} {\trimBT} {\trimLR} {\trimBT}},clip,width=\linewidth]{diagrams/parametric-case/parametric_resol_\x_inference_probs.pdf}
            \caption{Inference Probs., {\x}x{\x}}
        \end{subfigure}
        %
        \begin{subfigure}[b]{\subfigwidth\linewidth}
            \centering
            \includegraphics[trim={{\trimLR} {\trimBT} {\trimLR} {\trimBT}},clip,width=\linewidth]{diagrams/parametric-case/parametric_resol_\x_inference_bin.pdf}
            \caption{Inference Binary, {\x}x{\x}}
        \end{subfigure}
        \clearpage
    }
    \caption{Model is able to make accurate predictions on input point cloud image at multiple resolutions. For some applications, low resolution and high inference speed may be tolerable. For others, higher resolution is needed to accurate convey visible regions. All models demonstrate resilience to adversaries at each level.}
    \label{fig:parametric-resolution}
\end{figure*}
\renewcommand{\trimLR}{0}
\renewcommand{\trimBT}{0}
\renewcommand{\subfigwidth}{0.9}
\begin{figure}[t]
    \centering
    \begin{subfigure}[b]{\subfigwidth\linewidth}
        \centering
        \includegraphics[trim={{\trimLR} {\trimBT} {\trimLR} {\trimBT}},clip,width=\linewidth]{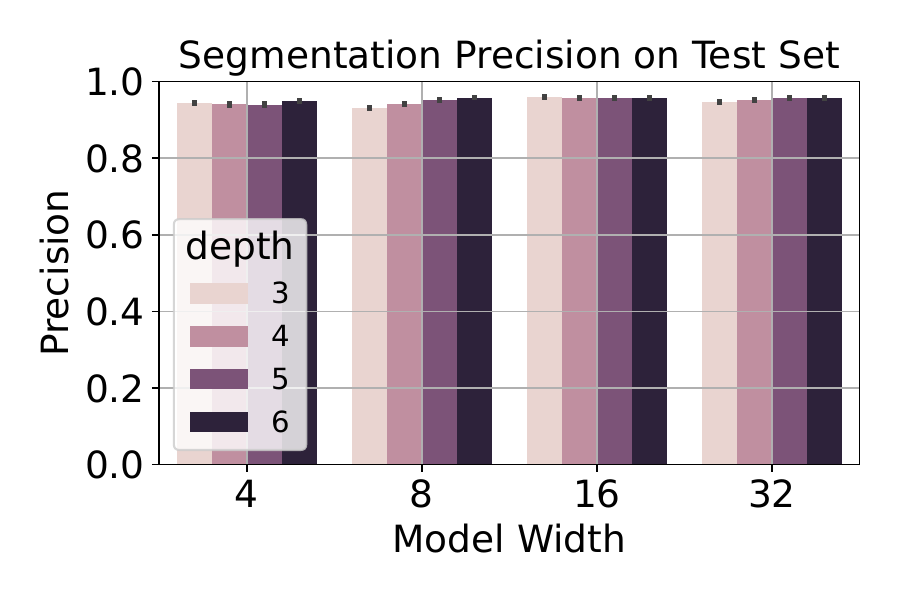}
        \caption{Segmentation precision at depths and widths.}
        \label{fig:parametric-performance-prec}
    \end{subfigure}
    \begin{subfigure}[b]{\subfigwidth\linewidth}
        \centering
        \includegraphics[trim={{\trimLR} {\trimBT} {\trimLR} {\trimBT}},clip,width=\linewidth]{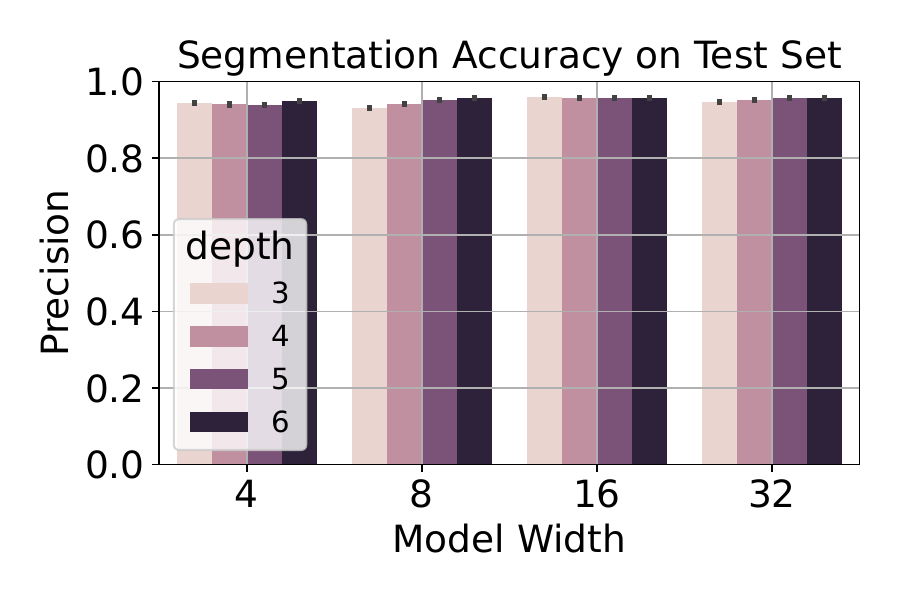}
        \caption{Segmentation accuracy by depths and widths.}
        \label{fig:parametric-performance-acc}
    \end{subfigure}    
    \caption{Model performance is consistent across parameter selection with slight increases with increased model width. We used early stopping criteria to prevent overfitting. Parametric understanding of model performance informs selection of parameters in variety of environments.}
    \label{fig:parametric-performance}
\end{figure}
    

We evaluate models with width ranging from 8 to 64 channels, depth ranging from 3 to 6 layers, and input resolution ranging from squares with side lengths of 64 to 512 pixels that discretize the 2D BEV domain. Figure~\ref{fig:parametric-performance} describes the performance across width and model depth. The model performs similarly across all combinations, aided by early stopping criteria to prevent overfitting. This high degree of performance suggests an easy ability to adapt the model to circumstances with compute and memory restrictions. 

Since the model's compute time scales superlinearly with input size, it is important to identify when lower resolution is sufficient. Figure~\ref{fig:parametric-resolution} illustrates differences in outcomes for point cloud processing of a scene from the CARLA dataset at resolutions ranging from 64 to 512 pixels per side. For a maximum point cloud range of $75~m$ in any direction, pixels therefore fall between accounting for between $2.5~m$ and $0.3~m$ per pixel. The degree of satisfaction with model performance depends on the application requirements.

Our parametric study ensures that the model remains adaptable to a broad range of real-world scenarios, from low-latency robotic perception to high-precision offline mapping. Additional discussion on application-related constraints is included in Appendix~\ref{appendix:applications}. 

\subsection{Real-Time Multi-Agent Integration}

Given its computational efficiency, FOV estimation can be integrated into real-time multi-agent robotic systems. We deploy our models in a ROS-based infrastructure, testing their performance on simulated agents operating in both ground-based and elevated infrastructure settings.

\begin{figure}[t]
    \centering
    \begin{subfigure}[b]{0.48\linewidth}
        \centering
        \includegraphics[width=0.95\linewidth,trim={5cm 4cm 10cm 7cm},clip,fbox]{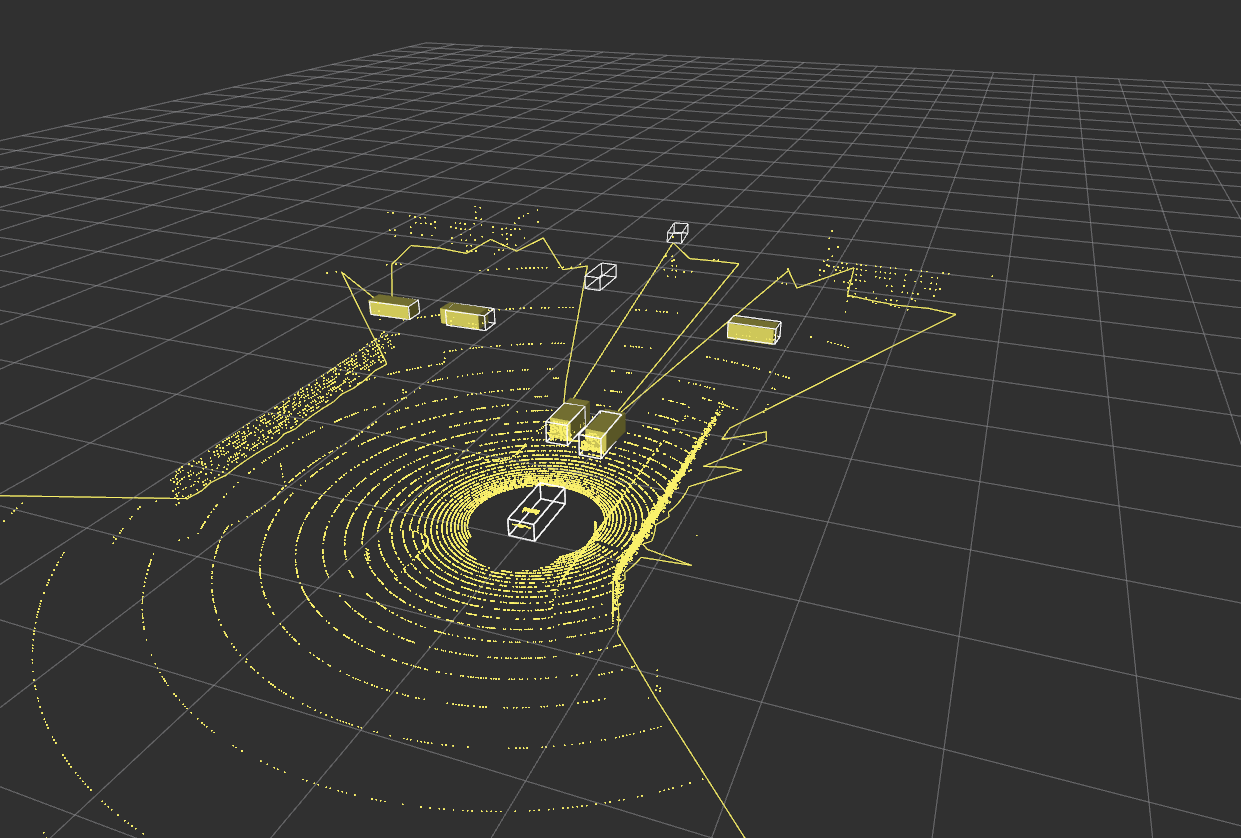}
        \caption{Ground-based agent.}
    \end{subfigure}
    %
    \begin{subfigure}[b]{0.48\linewidth}
        \centering
        \includegraphics[width=0.95\linewidth,trim={5cm 6cm 10cm 5cm},clip,fbox]{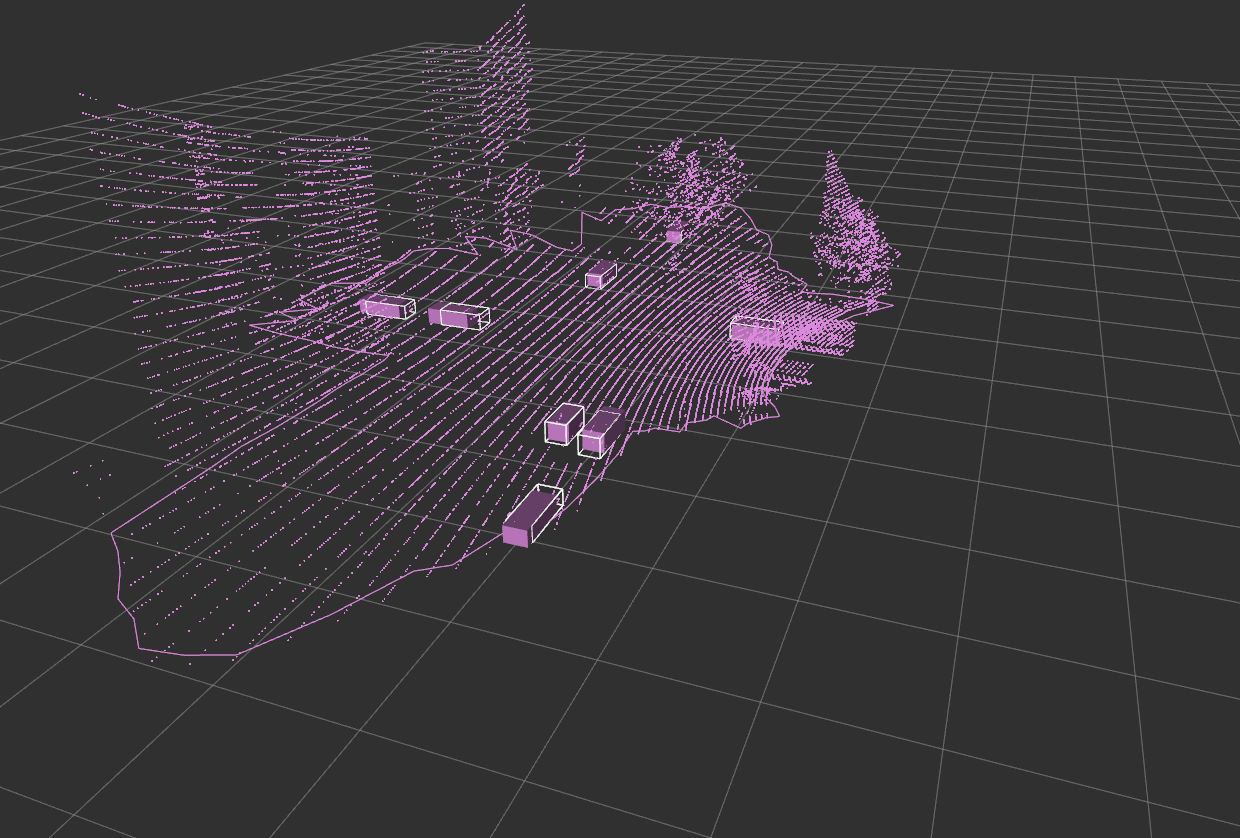}
        \caption{Infrastructure LiDAR.}
    \end{subfigure}
    %
    \begin{subfigure}[b]{0.48\linewidth}
        \centering
        \includegraphics[width=0.95\linewidth,trim={1cm 1cm 1cm 5cm},clip,fbox]{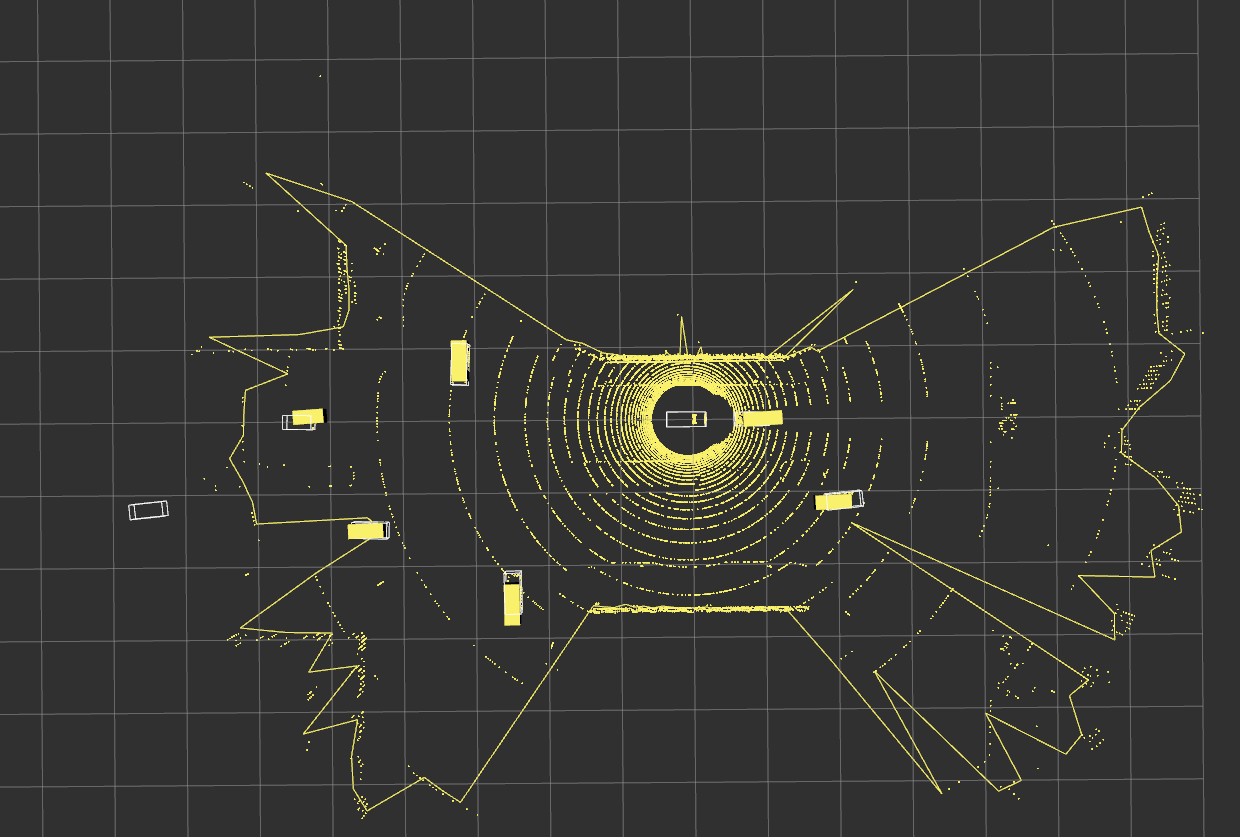}
        \caption{Ground-based agent, BEV.}
    \end{subfigure}
    %
    \begin{subfigure}[b]{0.48\linewidth}
        \centering
        \includegraphics[width=0.95\linewidth,trim={1cm 2cm 6.25cm 7cm},clip,fbox]{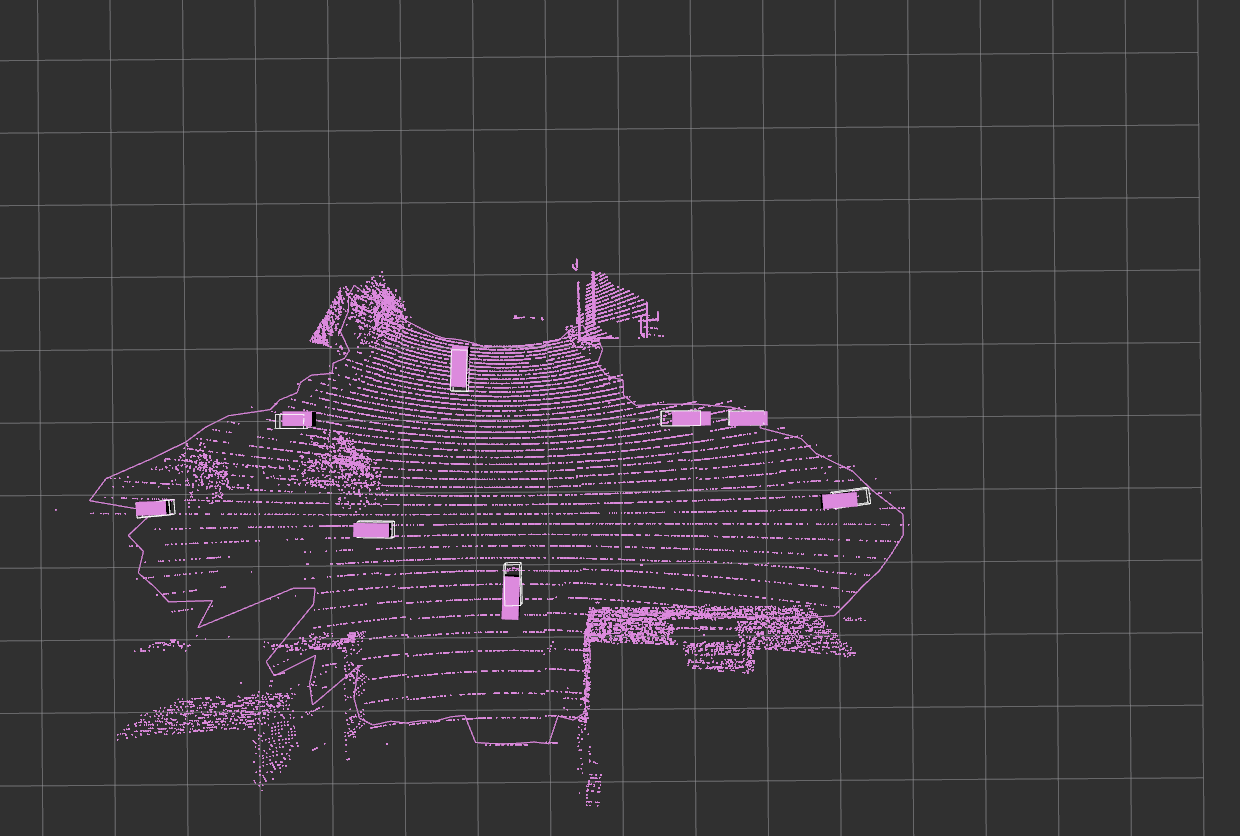}
        \caption{Infrastructure LiDAR, BEV.}
    \end{subfigure}
    \caption{FOV estimates integrated in real-time multi-agent robotic sensor fusion pipeline to perform information sharing and consistency checking among agents. (a) Ground-based agent shows object-based occlusions where field of view behind objects is limited. (b) Infrastructure sensor is not subject to object-level occlusions due to elevated vantage point and ability to see over objects.}
    \label{fig:results-ros-visualization}
\end{figure}

We use CARLA-derived datasets converted into ROS bags, enabling replay at adjustable frame rates. ROS nodes for perception, tracking, sensor fusion, and FOV estimation operate in real-time, receiving sensor data from a central publisher. The FOV estimation pipeline, including all preprocessing steps, achieves inference exceeding 50~Hz, making it viable for real-world deployment. Figure~\ref{fig:results-ros-visualization} visualizes the raw LiDAR point cloud, detected bounding boxes, and estimated FOV boundaries for a ground agent (yellow) and an elevated infrastructure sensor (purple) in two test scenes. The segmentation approach enables efficient FOV estimation with minimal computational overhead, enabling ground-based and infrastructure-mounted sensors can share estimated FOVs, improving collaborative situational awareness.

Visualizations, including dynamic multi-agent ROS-based testing, are provided in supplemental materials online.

\section{Conclusion}

FOV estimation is a critical yet underexplored component of secure sensor fusion in AVs. In this work, we introduced the first classical FOV estimators adapted from computer graphics and used them to construct the first labeled FOV datasets for AV perception. However, our analysis revealed that traditional FOV estimation techniques are highly vulnerable to adversarial attacks, such as LiDAR spoofing, making them unsuitable for contested environments. To address these vulnerabilities, we developed a deep learning-based FOV estimator trained on both benign and adversarial data, significantly improving robustness against attacks on sensing. Furthermore, we integrated Monte Carlo dropout to provide uncertainty-aware FOV predictions, enabling real-time anomaly detection. Through extensive experiments, we demonstrated strong generalization of our model across multiple datasets and adversarial conditions, ensuring feasibility for real-time deployment.

\section*{Statements}

\noindent \textbf{Ethics considerations.} This work proposes defenses against attacks on AVs and raises minimal ethical considerations. All research was conducted using datasets and simulations.

\vspace{8pt}
\noindent \textbf{Open science.}  As part of the compliance to the open science initiative, we will release all datasets, models, and source code used to derive the results in this work. The source code will be released online for researchers to reproduce our outcomes.

\clearpage
\bibliographystyle{plain}
\bibliography{references}

\appendix
\section{Model Training} \label{appendix:model-training}

\begin{figure}[th]
    \centering
    \includegraphics[width=0.8\linewidth]{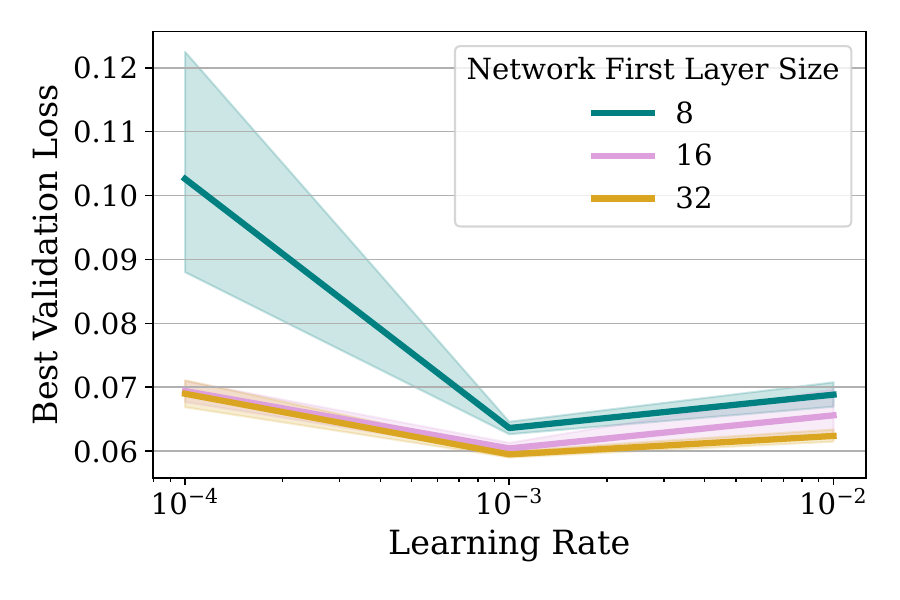}
    \caption{Learning rate converges for all network sizes using 5-fold crossvalidation. Crossvalidation also aids in selection of network size and dropout rate.}
    \label{fig:algorithm-kfold}
\end{figure}

We perform a simple crossvalidation procedure to select parameters for the model for testing and graphics visualization. A snapshot of the crossvalidation outcomes are included in Figure~\ref{fig:algorithm-kfold}. Parameters tested are included in Table~\ref{tab:algorithm-crossvalid}.

\begin{table}[t]
    \centering
    \begin{tabular}{c|c}
         Parameter & Input Domain \\
         \toprule
         Channel scaling factor & $[4, 8, 16, 32]$  \\
         Dropout rate & $[0.05, 0.10, 0.15]$ \\
         Learning rate  & $[0.0001, 0.001, 0.01]$ \\
    \end{tabular}
    \caption{Cross validation input parameters}
    \label{tab:algorithm-crossvalid}
\end{table}
\section{Application-Specific Considerations} \label{appendix:applications}

Applications for autonomy have different constraints and operating requirements. These applications will inform the requirements for execution rates and performance targets. We briefly discuss a selection of these applications here.

\begin{itemize}
    \item \textbf{Real-Time Robotics (AVs, Drones, UGVs):} Autonomous vehicles and mobile robots operate under strict latency requirements, often demanding inference rates exceeding 10–20 Hz. For these applications, lightweight models with optimized architectures are essential to maintain real-time performance while ensuring reliable perception in dynamic environments.
    \item \textbf{Infrastructure-Based Sensing:} Smart city deployments and fixed sensor networks (e.g., traffic monitoring, security surveillance) prioritize accuracy over inference speed, as they operate in a non-time-critical manner. Higher model depth and width may be preferable in these settings to enhance detection robustness.
    \item \textbf{High-Fidelity Mapping and Surveying:} Applications such as SLAM (Simultaneous Localization and Mapping) and large-scale environmental modeling benefit from high-resolution input data and deeper models to improve spatial precision. Computational efficiency remains a secondary concern compared to accuracy and completeness of the estimated FOV.
    \item \textbf{Contested Environments:} Defense and security-oriented autonomous systems must operate under adversarial conditions where sensor spoofing or occlusion attacks are likely. In such cases, models incorporating Monte Carlo dropout (MCD) and adversarial training are essential, even at the cost of higher computational demands.
\end{itemize}

\end{document}